\documentclass[11pt]{article}

\usepackage[final]{acl}
\usepackage{times}
\usepackage{latexsym}
\usepackage[T1]{fontenc}
\usepackage[utf8]{inputenc}
\usepackage{microtype}
\usepackage{graphicx}
\usepackage{amsmath}
\usepackage{amssymb}
\usepackage{booktabs}
\usepackage{multirow}
\usepackage{xcolor}
\usepackage{enumitem}
\usepackage{pifont}
\usepackage{algorithm}
\usepackage{algpseudocode}
\usepackage{tikz}
\usetikzlibrary{positioning,arrows.meta,fit,backgrounds,calc,shapes.geometric}

\definecolor{trblue}{HTML}{2A78D6}
\definecolor{traqua}{HTML}{1BAF7A}
\definecolor{trorange}{HTML}{EB6834}
\definecolor{trviolet}{HTML}{7A5CC4}
\definecolor{trgray}{HTML}{6B6A66}

\newcommand{\TR}{TableRouter}
\newcommand{\cmark}{\ding{51}}
\newcommand{\xmark}{\ding{55}}

\setlist{leftmargin=*,itemsep=1pt,topsep=3pt,parsep=0pt}
\title{TableRouter: Query-Aware Evidence Construction\\for Long-Table Question Answering}

\author{
\\
 \textbf{Anahita Bolourani},
 \textbf{Xiaofeng Lin},
 \textbf{Tung Sum Thomas Kwok},
 \textbf{Guang Cheng}
\\
\\
 University of California, Los Angeles
\\
 \small{
   \textbf{Correspondence:} \href{mailto:anabolourani@ucla.edu}{anabolourani@ucla.edu}
 }}

\begin{document}
\maketitle

\begin{abstract}
LLMs can reason over tables, but when tables outgrow the context budget, retrieving relevant rows is only part of the solution. A retrieved subset may omit information required for an exact aggregate; even complete evidence may remain difficult for an LLM to use. We introduce \TR{}, a training-free framework that analyzes each query, then chooses between two routes: a computation route, which derives question-conditioned statistics from the full table, and a coverage route, which retrieves relevant rows and column summaries. The chosen context is passed to a frozen LLM in one call. Across three LLMs and three large-table benchmarks, \TR{} outperforms the strongest of six single-call baselines in all nine settings by 1.2--18.3 accuracy points, using 77--95\% fewer estimated FLOPs than truncated-table prompting, with indexing amortized across questions. The gains extend to long tables from databases not used in earlier experiments, where \TR{} achieves 35.8\% with Qwen2.5-7B, versus 16.4\% for the strongest external baseline. A controlled experiment shows that explicitly providing computed statistics improves accuracy by up to 40.0 points, even when all answer-determining rows are available. These results distinguish evidence access from evidence use: long-table reasoning depends on how information is represented, not only on which rows are retrieved.
\end{abstract}

\begin{figure}[t]
\centering
\includegraphics[width=\columnwidth]{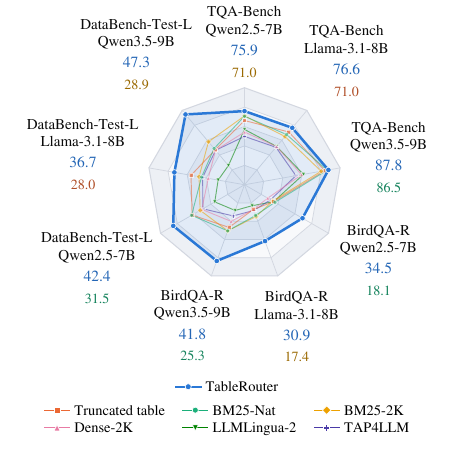}
\caption{Observed accuracy across three LLMs and three benchmarks. \TR{} leads the six primary external baselines in each setting. Radial ranges are 0--100\% for TQA-Bench and 0--50\% otherwise. Under each label: \TR{}, then the strongest baseline in its color. Values appear in Table~\ref{tab:main}; internal context policies are compared separately in Table~\ref{tab:internal}.}
\label{fig:radar}
\end{figure}

\section{Introduction}
\label{sec:intro}

A short table context can fail in two different ways: it can omit information needed to determine the answer, or it can preserve that information in a form the language model does not successfully use. These correspond to two distinct requirements: \emph{evidence access} and \emph{evidence use}. Reducing a table to relevant rows addresses input length, but does not necessarily satisfy either requirement.

Consider the question, ``What is the average price of Toyota cars made after 2015?'' In the illustrative table of Figure~\ref{fig:example}, 1,284 of 48,213 listings satisfy the conditions. A 2,000-token context holds only a fraction of them. Even perfect relevance ranking supplies a sample rather than the full population: its mean need not equal the requested mean. Computing the filtered statistic over the full table expresses the answer-determining information in one line. By contrast, looking up the price of listing 31907 calls for the original identifying cells, not a table-wide summary.

\begin{figure}[t]
\centering
\footnotesize\setlength{\fboxsep}{2.2pt}
\fcolorbox{black!30}{white}{%
\begin{minipage}{0.96\columnwidth}
\textbf{Aggregation} on a table of 48,213 car listings:\\
\emph{What is the average price of Toyota cars made after 2015?}
\end{minipage}}\\[3pt]
\fcolorbox{trgray!70}{trgray!7}{%
\begin{minipage}{0.96\columnwidth}
\textbf{Rows retrieved by similarity} (2,000 tokens)\\[2pt]
{\ttfamily\fontsize{6.6}{8}\selectfont
| make~~~| model~~~| year | price |\\
| Toyota | Corolla~| 2017 | 18900 |\\
\dots{} 130 more rows}\\[2pt]
LLM: \emph{21,370}, the mean of the rows it sees \textcolor{trorange}{\xmark}
\end{minipage}}\\[3pt]
\fcolorbox{traqua!70!black}{traqua!7}{%
\begin{minipage}{0.96\columnwidth}
\textbf{\TR{}, computation route} (excerpt)\\[2pt]
{\ttfamily\fontsize{6.6}{8}\selectfont
Computed over the full table (48,213 rows):\\
make = Toyota and year > 2015:~~1,284 rows (2.7\%)\\
price | make = Toyota and year > 2015:\\
\hspace*{1em}count 1,284, mean 24,890.0, min 6,500, max 71,200\\
\dots{} then the rows behind min and max}\\[2pt]
LLM: \emph{24,890} \textcolor{traqua!70!black}{\cmark}
\end{minipage}}\\[3pt]
\fcolorbox{black!30}{white}{%
\begin{minipage}{0.96\columnwidth}
\textbf{Lookup} on the same table:\\
\emph{What is the price of car listing 31907?}
\end{minipage}}\\[3pt]
\fcolorbox{trblue!70!black}{trblue!6}{%
\begin{minipage}{0.96\columnwidth}
\textbf{\TR{}, coverage route} (excerpt)\\[2pt]
{\ttfamily\fontsize{6.6}{8}\selectfont
| id~~~~| make~~~| model~~~| year | price |\\
| 31907 | Toyota | Corolla~| 2017 | 18900 |\\
\dots{} rows matching question values, then ranked rows}\\[2pt]
LLM: \emph{18,900} \textcolor{traqua!70!black}{\cmark}
\end{minipage}}
\caption{One table, two forms of evidence. An exact filtered
aggregate depends on the full qualifying population, whereas
a lookup requires identifying cells.
}
\label{fig:example}
\end{figure}

This distinction makes evidence construction more than a row-selection problem. A computed statistic can represent information distributed across many rows and reduce the arithmetic left to the LLM. Original rows preserve associations that an aggregate can erase. The appropriate evidence representation therefore depends on the question, the context budget, and the LLM receiving it.

We introduce \TR{}, a query-aware, training-free evidence-construction framework for a frozen LLM. Its \emph{computation route} computes question-conditioned statistics over the full supplied table, adding supporting rows when space permits. Its \emph{coverage route} selects relevant rows and column summaries under a context budget. A rule selects the computation route when the question contains a computational cue and at least one candidate statistic passes a lexical grounding check; otherwise, it uses the coverage route. If the entire table fits within the coverage route's budget, it is passed to the LLM without row selection. Both routes share a reusable table index, and neither requires a generative preprocessing call. 

Our evaluation addresses three questions.  Can compact, question-dependent evidence improve long-table QA accuracy while reducing computational cost? Can computed statistics improve accuracy even when all answer-determining rows are already available? How does choosing between computed statistics, retrieved rows, and their combination affect accuracy and computational cost?

Across three LLMs and three large-table benchmarks, \TR{} consistently outperforms the evaluated single-call baselines, while substantially reducing estimated compute relative to truncated-table prompting when indexing is amortized across questions. The advantage extends to a separate post-freeze evaluation on long tables from previously unused databases.

Beyond system-level performance, our experiments examine why evidence construction matters. A controlled study shows that explicitly providing computed statistics can improve accuracy even when all answer-determining rows are already available. Under a shared context budget, combining computed statistics with coverage-based evidence further improves performance across all three LLMs on BirdQA and DataBench.

Comparisons of fixed and routed contexts further highlight the complementary roles of different evidence representations: original rows preserve associations needed for lookups, while computed statistics summarize information distributed across many rows. These comparisons also show how the choice of representation shapes the trade-off between accuracy and computational cost.

Together, these results establish that \emph{what a long-table context makes explicit is a consequential design choice, separate from how many relevant rows it contains}.

This paper makes three contributions:
\begin{itemize}
    \item We introduce a training-free framework that combines full-table computation, coverage-based retrieval, and reusable indexing to construct evidence for a single frozen-LLM call.

    \item We demonstrate improvements across three LLMs and three benchmarks, with lower estimated compute than truncated-table prompting and additional post-freeze evaluation.

    \item Controlled experiments establish the value of explicit computations beyond row access, while policy comparisons characterize accuracy--compute trade-offs across evidence representations.
\end{itemize}

\section{Related Work}
\label{sec:related}

 TaPas and TaBERT learn table-aware representations, while TableLlama adapts an LLM to tabular tasks through instruction tuning \citep{herzig2020tapas,yin2020tabert,zhang2024tablellama}. Long-context and robustness studies document the sensitivity of table reasoning to input length, format, and structural variation \citep{wolff2025well,li2025longtablebench}. Unlike approaches that adapt the language model, \TR{} constructs question-dependent evidence for frozen LLMs without task-specific training.

Other approaches focus on selecting and compressing table evidence. ITR retrieves question-relevant subtables \citep{lin2023itr}; TableRAG uses LM-generated query expansion to retrieve schema and cell values \citep{chen2024tablerag}; and LLMLingua-2 compresses prompts with a trained encoder \citep{pan2024llmlingua2}. TAP4LLM combines table sampling, augmentation, and packing, including table-level statistical features \citep{sui2024tap4llm}. Recent approaches further address subtable selection and noisy table inputs through divide-and-conquer selection and denoising
\citep{lee2026piece,ye2026noise}. Retrieved rows may omit values needed for exact aggregation, while table-level summaries need not reflect the conditions specified by a question. \TR{} complements coverage-based selection with question-conditioned statistics computed over the full supplied table, rather than only the retrieved subset.

Explicit computation provides another approach to table reasoning. TaCube generates question-sensitive arithmetic and aggregation results and appends selected computations to table inputs \citep{zhou2022tacube}. In contrast, \TR{} constructs a bounded replacement context for long tables, using full-table conditional statistics or coverage-based retrieval according to the question. This design does not require task-specific training or generative preprocessing and supports frozen LLMs.

\section{\TR{}}
\label{sec:method}

\begin{figure*}[t]
\centering
\tikzset{
  pics/tabicon/.style={code={
    \fill[#1!30] (-0.32,0.12) rectangle (0.32,0.24);
    \draw[#1!80!black, line width=0.35pt] (-0.32,0.12) -- (0.32,0.12) (-0.32,0) -- (0.32,0)
      (-0.32,-0.12) -- (0.32,-0.12) (-0.107,-0.24) -- (-0.107,0.24) (0.107,-0.24) -- (0.107,0.24);
    \draw[#1!80!black, line width=0.6pt, rounded corners=0.6pt] (-0.32,-0.24) rectangle (0.32,0.24);
  }},
  pics/qicon/.style={code={
    \fill[#1] (-0.1,-0.17) -- (-0.22,-0.31) -- (0.0,-0.22) -- cycle;
    \draw[#1, line width=0.8pt, fill=white] (0,0) circle (0.23);
    \node[font=\bfseries\small, text=#1, inner sep=0pt] at (0,0) {?};
  }},
  pics/searchicon/.style={code={
    \draw[#1, line width=0.85pt] (-0.05,0.05) circle (0.15);
    \draw[#1, line width=1.5pt, line cap=round] (0.06,-0.06) -- (0.21,-0.21);
  }},
  pics/funnelicon/.style={code={
    \draw[#1, line width=0.75pt, line join=round, fill=#1!15]
      (-0.24,0.2) -- (0.24,0.2) -- (0.05,-0.03) -- (0.05,-0.24) -- (-0.05,-0.18) -- (-0.05,-0.03) -- cycle;
  }},
  pics/calcicon/.style={code={
    \draw[#1, line width=0.75pt, rounded corners=1.5pt, fill=white] (-0.18,-0.25) rectangle (0.18,0.25);
    \fill[#1!45] (-0.12,0.09) rectangle (0.12,0.19);
    \foreach \x in {-0.08,0,0.08} \foreach \y in {-0.17,-0.08,0.01} {\fill[#1] (\x,\y) circle (0.022);}
  }},
  pics/checkdoc/.style={code={
    \draw[#1, line width=0.7pt, line join=round, fill=white]
      (-0.18,-0.25) -- (-0.18,0.25) -- (0.08,0.25) -- (0.18,0.15) -- (0.18,-0.25) -- cycle;
    \draw[#1, line width=0.4pt] (-0.11,0.12) -- (0.06,0.12) (-0.11,0.03) -- (0.1,0.03) (-0.11,-0.06) -- (0.02,-0.06);
    \fill[#1] (0.17,-0.19) circle (0.11);
    \draw[white, line width=0.8pt, line cap=round, line join=round] (0.12,-0.19) -- (0.16,-0.23) -- (0.22,-0.15);
  }},
  pics/dbicon/.style={code={
    \draw[#1, line width=0.9pt, fill=white]
      (-0.35,-0.3) arc[start angle=180, end angle=360, x radius=0.35, y radius=0.1] -- (0.35,0.3)
      arc[start angle=0, end angle=180, x radius=0.35, y radius=0.1] -- cycle;
    \foreach \y in {0.3,0.1,-0.1} \draw[#1, line width=0.9pt] (-0.35,\y) arc[start angle=180, end angle=360, x radius=0.35, y radius=0.1];
  }},
  pics/stackicon/.style={code={
    \foreach \d in {0.12,0.06,0} {\draw[#1, line width=0.5pt, fill=white] (-0.22+\d,-0.2+\d) rectangle (0.12+\d,0.14+\d);}
    \draw[#1, line width=0.4pt] (-0.17,0.06) -- (0.07,0.06) (-0.17,-0.02) -- (0.07,-0.02) (-0.17,-0.1) -- (0.02,-0.1);
  }},
  pics/clustericon/.style={code={
    \draw[#1!70, dashed, line width=0.4pt] (-0.27,0) circle (0.16);
    \foreach \p/\c in {(-0.34,0.06)/trorange,(-0.22,0.08)/traqua,(-0.27,-0.03)/trblue,(-0.35,-0.07)/traqua,(-0.19,-0.07)/trorange}
      {\fill[\c] \p circle (0.026);}
    \draw[-{Stealth[length=1.1mm]}, #1, line width=0.5pt] (-0.09,0) -- (0.07,0);
    \draw[#1!70, dashed, line width=0.4pt] (0.27,0) circle (0.16);
    \foreach \p/\c in {(0.2,0.07)/trorange,(0.25,0.09)/trorange,(0.33,0.05)/traqua,(0.35,0.0)/traqua,(0.23,-0.07)/trblue,(0.29,-0.08)/trblue}
      {\fill[\c] \p circle (0.026);}
  }},
  pics/bmicon/.style={code={
    \draw[#1, line width=0.65pt, line join=round, fill=white]
      (-0.2,-0.25) -- (-0.2,0.25) -- (0.06,0.25) -- (0.16,0.15) -- (0.16,-0.25) -- cycle;
    \draw[#1, line width=0.4pt] (-0.13,0.12) -- (0.06,0.12) (-0.13,0.03) -- (0.09,0.03);
    \draw[#1, line width=0.8pt, fill=white] (0.08,-0.1) circle (0.1);
    \draw[#1, line width=1.3pt, line cap=round] (0.15,-0.17) -- (0.25,-0.27);
  }},
  pics/barsicon/.style={code={
    \fill[#1!55] (-0.2,-0.2) rectangle (-0.11,0.0);
    \fill[#1!80] (-0.06,-0.2) rectangle (0.03,0.12);
    \fill[#1!100] (0.08,-0.2) rectangle (0.17,0.22);
    \draw[#1!80!black, line width=0.5pt] (-0.24,-0.2) -- (0.21,-0.2);
  }},
}

\newcommand{\trlogo}[2]{%
  \IfFileExists{Figures/#1.pdf}{\includegraphics[height=0.62cm,width=1.8cm,keepaspectratio]{Figures/#1.pdf}}{%
  \IfFileExists{Figures/#1.png}{\includegraphics[height=0.62cm,width=1.8cm,keepaspectratio]{Figures/#1.png}}{%
  \fcolorbox{black!30}{white}{\rule[-0.13cm]{0pt}{0.45cm}\footnotesize\bfseries\color{black!70}#2}}}}

\newcommand{\icard}[7]{
  \node[draw=#2!70!black, fill=#2!7, rounded corners=4pt, line width=0.6pt,
        minimum width=#4, minimum height=0.5cm, inner sep=0pt, shadow] (#1) at #3 {};
  \pic[scale=0.75] at ([xshift=0.42cm]#1.west) {#5};
  \node[anchor=west, inner sep=0pt, font=\small\bfseries, text=#2!80!black] (#1t)
    at ([xshift=0.92cm]#1.west) {#6};
  \node[anchor=base west, inner sep=0pt, font=\footnotesize, text=black!70]
    at ([xshift=0.22cm]#1t.base east) {#7};
}

\newcommand{\stepcard}[6]{
  \node[draw=#2!70!black, fill=#2!7, rounded corners=4pt, line width=0.6pt,
        minimum width=2.4cm, minimum height=1.95cm, inner sep=0pt, shadow] (#1) at #3 {};
  \pic[scale=1.1] at ([yshift=-0.42cm]#1.north) {#4};
  \node[font=\small\bfseries, text=#2!80!black, align=center, anchor=north, inner sep=0pt]
    (#1t) at ([yshift=-0.8cm]#1.north) {#5};
  \node[font=\footnotesize, text=black!70, align=center, anchor=north, inner sep=0pt]
    at ([yshift=-0.09cm]#1t.south) {#6};
}

\newcommand{\routepanel}[5]{
  \node[fill=#2!6, rounded corners=5pt, minimum width=4.2cm, minimum height=1.74cm,
        inner sep=0pt, shadow] (#1) at #3 {};
  \fill[#2!85!black, rounded corners=5pt] ([yshift=-0.66cm]#1.north west) rectangle (#1.north east);
  \fill[#2!85!black] ([yshift=-0.66cm]#1.north west) rectangle ([yshift=-0.33cm]#1.north east);
  \draw[#2!75!black, line width=0.8pt, rounded corners=5pt] (#1.south west) rectangle (#1.north east);
  \node[text=white, align=center, inner sep=0pt] at ([yshift=-0.33cm]#1.north)
    {{\small\bfseries #4}\\[-2pt]{\scriptsize #5}};
}

\newcommand{\rchip}[4]{
  \node[draw=black!18, fill=white, rounded corners=3pt, line width=0.5pt,
        minimum width=3.84cm, minimum height=0.42cm, inner sep=0pt] (#1) at #2 {};
  \pic[scale=0.68] at ([xshift=0.3cm]#1.west) {#3};
  \node[anchor=west, inner sep=0pt, font=\footnotesize\bfseries, text=black!80]
    at ([xshift=0.62cm]#1.west) {#4};
}
\resizebox{\textwidth}{!}{%
\begin{tikzpicture}[
  font=\small,
  shadow/.style={preaction={fill=black!14, transform canvas={xshift=0.9pt, yshift=-0.9pt}}},
  arr/.style={-{Stealth[length=2.2mm,width=1.7mm]}, line width=0.8pt, draw=black!65, rounded corners=4pt},
  carr/.style={arr, draw=traqua!85!black},
  varr/.style={arr, draw=trblue!85!black},
  rarr/.style={arr, draw=trviolet!80!black},
  darr/.style={-{Stealth[length=2.0mm,width=1.5mm]}, line width=0.75pt, draw=trviolet!85!black, dashed, rounded corners=4pt},
  halo/.style={preaction={draw=#1, line width=3.6pt, -}},
  pill/.style={font=\scriptsize\bfseries, text=white, fill=#1!85!black, rounded corners=2pt,
               inner xsep=2.5pt, inner ysep=1.3pt},
  tag/.style={font=\small\bfseries, text=white, fill=#1!85!black, rounded corners=6pt,
              inner xsep=7pt, inner ysep=3pt},
  note/.style={font=\footnotesize\itshape, inner sep=0pt},
]

\begin{scope}[on background layer]
  \draw[trviolet!45, fill=trviolet!5, line width=0.6pt, rounded corners=8pt] (-0.3,5.95) rectangle (23.2,7.5);
  \draw[trorange!45, fill=trorange!4, line width=0.6pt, rounded corners=8pt] (-0.3,0.0) rectangle (23.2,4.95);
\end{scope}
\node[tag=trviolet, anchor=east] at (23.05,7.5) {1\enspace Offline, once per table};
\node[tag=trorange, anchor=east] at (23.05,4.95) {2\enspace Per question};

\pic[scale=1.3] at (0.9,6.95) {tabicon=trgray};
\node[font=\small\bfseries, inner sep=0pt] at (0.9,6.38) {Table $T$};
\icard{enc}{trviolet}{(5.15,7.02)}{5.5cm}{stackicon=trviolet}{Row encoding}{BGE-small}
\icard{skm}{trviolet}{(11.6,7.02)}{6.6cm}{clustericon=trviolet}{Spherical $k$-means}{prototypes $p_k$}
\icard{lex}{trviolet}{(6.55,6.38)}{8.3cm}{bmicon=trviolet}{Lexical index}{BM25, value postings, rare cells}
\draw[arr] (1.38,6.95) -- (1.9,6.95) |- (enc.west);
\draw[arr] (1.9,6.95) |- (lex.west);
\draw[arr] (enc) -- (skm);
\draw[arr] (skm.east) -| (15.35,6.7) -- (15.65,6.7);
\draw[arr] (lex.east) -| (15.35,6.7);
\pic[scale=1.25] at (16.15,6.7) {dbicon=trviolet};
\node[anchor=west, inner sep=0pt, font=\small\bfseries, text=trviolet!80!black] at (16.75,6.88) {Reusable index};
\node[anchor=west, inner sep=0pt, font=\footnotesize, text=black!70] at (16.75,6.5) {reused for every question};

\pic[scale=1.5] at (0.9,3.2) {qicon=trgray};
\node[font=\small\bfseries, inner sep=0pt] at (0.9,2.52) {Question $q$};
\stepcard{mt}{trviolet}{(3.3,3.03)}{searchicon=trviolet}{Query\\matching}{value anchors}
\stepcard{cd}{trorange}{(6.05,3.03)}{funnelicon=trorange}{Conditions\\\& targets}{lexical rules}
\stepcard{fc}{traqua}{(8.8,3.03)}{calcicon=traqua}{Full-table\\computation}{over \emph{all} rows}
\stepcard{gr}{trorange}{(11.55,3.03)}{checkdoc=trorange}{Lexical\\grounding}{facts $G(q,T)$}
\node[diamond, aspect=1.6, draw=trorange!85!black, fill=trorange!12, line width=0.7pt,
      align=center, inner sep=0pt, font=\footnotesize\bfseries, shadow] (gate) at (14.15,3.03)
  {Computed\\evidence?};
\draw[arr] (1.32,3.03) -- (mt.west);
\draw[arr] (mt) -- (cd);
\draw[arr] (cd) -- (fc);
\draw[arr] (fc) -- (gr);
\draw[arr] (gr) -- (gate);

\routepanel{comp}{traqua}{(18.1,3.61)}{Computation route}{$\le$\,600 tokens}
\rchip{st}{(18.1,3.53)}{barsicon=traqua}{Statistics}
\rchip{sr}{(18.1,3.03)}{tabicon=traqua}{Supporting rows}
\routepanel{cov}{trblue}{(18.1,1.37)}{Coverage route}{$\le$\,2,000 tokens}
\rchip{sel}{(18.1,1.29)}{tabicon=trblue}{Selected rows}
\rchip{sum}{(18.1,0.79)}{barsicon=trblue}{Column summaries}
\draw[carr] (gate.north) |- node[pos=0.72, pill=traqua] {Yes} (comp.west |- 0,4.15);
\draw[varr] (gate.south) |- node[pos=0.72, pill=trblue] {No} (cov.west |- 0,1.91);

\node[draw=trgray!80!black, fill=trgray!7, rounded corners=4pt, line width=0.6pt,
      minimum width=2.2cm, minimum height=2.4cm, inner sep=0pt, shadow] (llm) at (21.8,2.45) {};
\node[font=\Large\bfseries, text=trgray!85!black, inner sep=0pt] at ([yshift=-0.38cm]llm.north) {LLM};
\node[inner sep=0pt] at ([yshift=0.12cm]llm.center) {\trlogo{llama-logo}{Llama}};
\node[inner sep=0pt] at ([yshift=-0.68cm]llm.center) {\trlogo{qwen-logo}{Qwen}};
\node[draw=trgray!80!black, fill=white, rounded corners=0.27cm, minimum width=1.9cm, minimum height=0.55cm,
      font=\small\bfseries, shadow] (ans) at (21.8,0.52) {Answer};
\draw[carr] (comp.east |- 0,3.3) -- (llm.west |- 0,3.3);
\draw[varr] (cov.east |- 0,1.6) -- (llm.west |- 0,1.6);
\draw[arr] (llm) -- (ans);

\draw[rarr] ([xshift=0.45cm]mt.north) |- (18.1,4.72) -- (comp.north);
\draw[rarr] (mt.south) |- (18.1,0.25) -- (cov.south);
\node[note, text=trviolet!80!black, anchor=south] at (8.8,0.33) {ranked rows $p_i$, used by both routes};

\draw[darr] (16.15,6.28) -- (16.15,5.25) -| ([xshift=-0.45cm]mt.north);
\node[note, text=trviolet!80!black, anchor=west] at (16.3,5.68) {index lookups};
\draw[arr, line width=1.1pt, halo=white] (0.9,6.2) -- (0.9,5.5) -| (fc.north);
\node[note, font=\footnotesize\bfseries\itshape, text=black!70, anchor=south] at (4.6,5.56) {all rows of $T$};
\end{tikzpicture}}
\caption{\TR{} workflow. Full-table statistics and ranked rows
support query-dependent evidence construction. The gate $g$
(Eq.~\ref{eq:ground}) selects the context passed to a frozen
LLM in a single call.}
\label{fig:workflow}
\end{figure*}
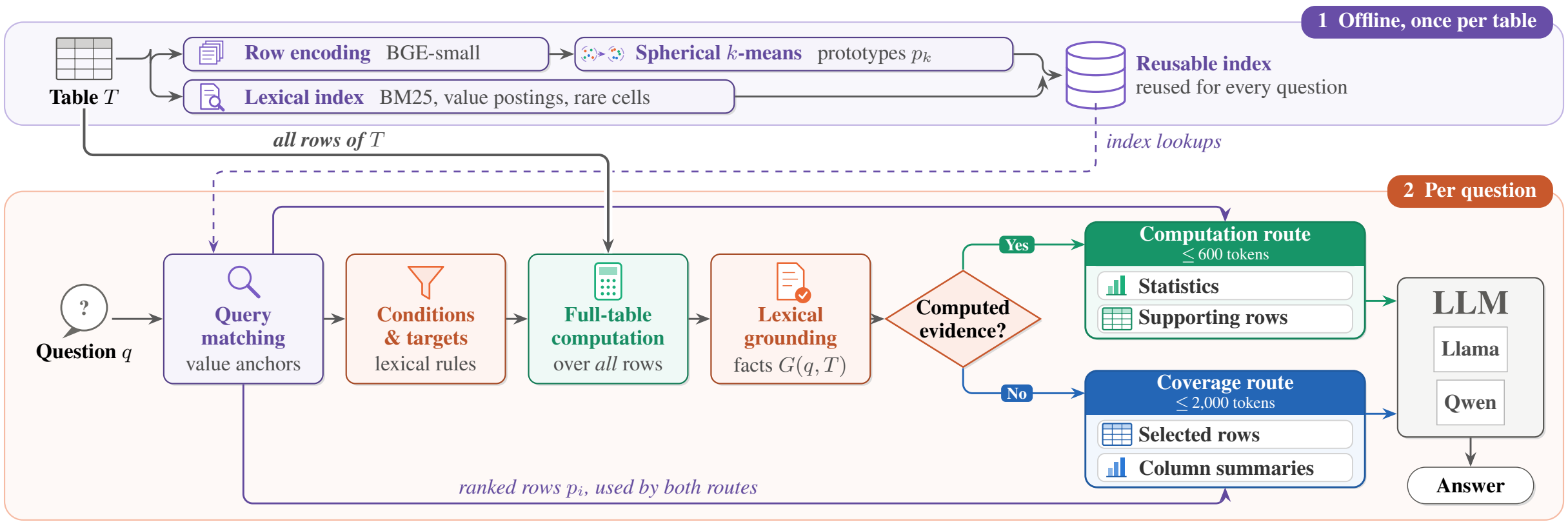

Given a question $q$ and a supplied table $T$ with $n$ rows and $m$ columns, \TR{} builds a compact evidence context $X(q,T)$ for a frozen LLM (Figure~\ref{fig:workflow}). It first indexes the table for reuse across questions. For each question, it then retrieves relevant rows, computes statistics from the full table, and selects between a \emph{computation context} (statistics and supporting rows) and a \emph{coverage context} (selected rows and column summaries). The LLM receives the chosen context in one call, without task-specific training or generative preprocessing.

\subsection{Building a Reusable Table Index}
\label{sec:index}

The index provides three complementary ways to locate table content: embedding similarity, lexical matching, and exact-value lookup.

Each table row is serialized with its column names and type-normalized values, then encoded by pretrained BGE-small \citep{xiao2024cpack} into a unit vector $x_i$, using at most 512 input tokens. Spherical $k$-means (SKM) groups the unit row vectors into $K$ clusters by cosine similarity \citep{dhillon2001concept}. Each cluster is represented by a normalized centroid $p_k$, called a \emph{prototype}. Each row has a primary cluster label $\ell(i)$ and a secondary label $\ell_2(i)$ used during candidate retrieval. 

The index also stores BM25 for word-based retrieval \citep{robertson2009bm25} and \emph{value postings}, which directly map each normalized cell value to the rows containing it. For example, a posting for \texttt{Toyota} identifies all rows with that value without relying on embedding similarity. The index also records \emph{rare cells}, values appearing at most twice, to help
recover distinctive matches.

These components serve complementary roles: SKM organizes row embeddings into clusters, whose prototypes identify semantically relevant groups; BM25 narrows candidates through lexical matching; and value postings retrieve exact matches, with rare values helping recover distinctive rows. The index is built once per table, allowing its construction cost to be shared across questions.

\subsection{Matching the Question and Ranking Rows}
\label{sec:retrieval}

The same encoder produces a unit question vector $u$. We also identify \emph{anchors}: values mentioned in the question that match table cells, such as names, dates, or numbers. Value postings locate the matching rows. An anchor is \emph{selective} when it matches at most $\max(5,0.2n)$ rows, and rarer anchors receive greater weight.

To identify relevant row groups, we compare the question vector $u$ with the SKM prototypes. For cluster $C_k=\{i:\ell(i)=k\}$, we compute a relevance score $s_k$ and normalize these scores into prototype weights $\alpha_k$:
\begin{equation}
\begin{aligned}
s_k &= p_k^\top u/0.1-0.05\ln(1+|C_k|),\\
\alpha_k &= \frac{\exp(s_k)}{\sum_{h=1}^{K}\exp(s_h)}.
\end{aligned}
\label{eq:prototype_weights}
\end{equation}
Here $p_k^\top u$ measures semantic similarity, while the second term penalizes large clusters. The normalized weights $\alpha_k$ determine which clusters are considered for row retrieval. Prototypes are taken in decreasing weight until their cumulative weight reaches 0.95. Rows whose primary or secondary labels belong to these prototypes form the initial candidate pool.

BM25 restricts the pool for large tables, while rare-value and selective-anchor matches can add rows. Aggregation-cue questions and tables with fewer than 30 rows use all rows as retrieval candidates. Thresholds and fallbacks are in Appendix~\ref{app:impl}.

For the resulting candidates $\mathcal C$, we combine
embedding similarity, normalized BM25 scores $\widetilde b_i$,
and weighted anchor coverage $\eta_i$:
\begin{equation}
\begin{aligned}
r_i &= x_i^\top u+0.5\,\widetilde b_i+\eta_i,\\
p_i &= \frac{\alpha_{\ell(i)}\exp(r_i/0.1)}
{\sum_{h\in\mathcal{C}}\alpha_{\ell(h)}\exp(r_h/0.1)}.
\end{aligned}
\label{eq:posterior}
\end{equation}
Thus $p_i$ combines row-level matches with the relevance of the row's primary cluster.

Columns are ranked separately using header, lexical, type, and value matches; named and identifier columns are retained; further rules are given in Appendix~\ref{app:impl}.

\subsection{Full-Table Statistics and Routing}
\label{sec:facts}

Lexical rules identify the requested operation, filtering conditions, and target columns, preferring columns named in the question. For each extracted condition and target, \TR{} computes statistics over all matching rows of the source table, not merely retrieved candidates $\mathcal C$. Supported results include counts, sums, means, extrema, and other question-dependent statistics. They are serialized as \emph{fact lines}, each potentially containing multiple statistics. Only candidate lines that pass later selection can appear in the LLM context. Supported operations, extraction limits, and fact-generation rules are detailed in Appendix~\ref{app:impl}.

Before routing, \emph{lexical grounding} checks whether these computed facts match column mentions or supported counting patterns in the question. Let $G(q,T)$ contain the grounded candidates, and let $\mathrm{comp}(q)$ detect computational cues, comparisons, or conditional yes/no questions. The bypass $b(T)$ is true for tables with at most 60 rows that fit within 600 tokens. The gate is
\begin{equation}
g(q,T)=\neg b(T)\wedge\mathrm{comp}(q)
\wedge\bigl(|G(q,T)|>0\bigr).
\label{eq:ground}
\end{equation}
It selects the evidence context as
\begin{equation}
X(q,T)=
\begin{cases}
X_{\mathrm{comp}}(q,T;B_{\mathrm{comp}}), & g(q,T)=1,\\
X_{\mathrm{cov}}(q,T;B_{\mathrm{cov}}), & \text{otherwise}.
\end{cases}
\label{eq:route}
\end{equation}
The route budgets are $B_{\mathrm{comp}}=600$ and
$B_{\mathrm{cov}}=2{,}000$ tokens, fixed across datasets.

\subsection{Evidence Context Construction}
\label{sec:contexts}
\label{sec:coverage}

\paragraph{Computation context.}

The computation context starts with ``Computed over the full table:'' and prioritizes condition-specific fact lines; remaining space may hold supporting rows for extrema, links, anchors, and high retrieval scores. Packing removes rows first, then ungrounded facts, then grounded facts that do not fit. Because routing precedes packing, some candidates considered by the gate may be omitted; every reported computation-route context retains at least one grounded fact.

\paragraph{Coverage context.}
If the table fits within the coverage budget $B_{\mathrm{cov}}$, it is passed through unchanged. Otherwise, \TR{} selects rows and column summaries to cover a set of \emph{information needs} $\mathcal J$ from the question. Each need $\nu\in\mathcal J$ represents a selective anchor or requested column operation and has an importance weight $w_\nu$.

The available evidence $\mathcal E$ consists of candidate rows from $\mathcal C$, individual numeric-column summaries, and a block of all summaries. These summaries describe entire columns, rather than question-filtered subsets. Each unit $e\in \mathcal E$ costs $c(e)$ tokens and provides coverage
$a_\nu(e)\in[0,1]$ for need $\nu$. We select a set
$S\subseteq\mathcal E$ to maximize
\begin{equation}
\begin{aligned}
F(S)&=\sum_{\nu\in\mathcal J}w_\nu
\min\Bigl(1,\sum_{e\in S}a_\nu(e)\Bigr),\\
\text{s.t.}\quad &c_0+\sum_{e\in S}c(e)\le B_{\mathrm{cov}},
\end{aligned}
\label{eq:objective}
\end{equation}
where $c_0$ accounts for the fixed table-header tokens in the constructed context. Capping each need's coverage rewards complementary evidence over repeated matches. The objective is monotone submodular \citep{lin2011submodular}.

A greedy heuristic selects affordable units with the largest positive coverage gain per token, then fills remaining space with ranked rows. Selected summaries precede a Markdown table of rows in their original order. Coverage does not guarantee answerability, and no approximation guarantee is claimed for this heuristic. Implementation details are in Appendix~\ref{app:impl}.


\section{Experimental Setup}
\label{sec:setup}

\paragraph{Datasets and evaluation status.} We evaluate three main benchmarks and a separate post-freeze BIRD set. TQA-Bench 16K/32K contains 1,200 four-option entity lookups on 120 tables, with related tables stacked identically for every method \citep{qiu2024tqabench}. BirdQA-R adapts 304 single-table questions from BIRD across 52 tables, capped at 200,000 rows, with gold answers recomputed on the supplied tables \citep{li2023bird}. DataBench-Test-L is the long-table subset of the SemEval-2025 DataBench test split, comprising 311 questions on nine tables exceeding a predeclared serialization threshold; all tables are used in full \citep{oses2025semeval}. We use TQA-Bench 64K and DataBench-Dev for method development \citep{oses2024databench}.

Post-freeze BIRD contains 299 questions on 55 tables (10,000--186,697 rows each) from 27 databases. Its questions were committed after the method freeze and before indexing or LLM calls. Earlier development used TQA-Bench and BirdQA-R test outputs; DataBench test inputs and some labels underwent LLM-free processing. We therefore distinguish the development-informed main comparison from the post-freeze evaluation. Appendix~\ref{app:protocol} records the chronology, exclusions, and later registrations.

\paragraph{LLMs and shared protocol.} We evaluate Qwen2.5-7B-Instruct, Llama-3.1-8B-Instruct, and Qwen3.5-9B in non-thinking mode \citep{qwen2025qwen25,grattafiori2024llama3,qwen2026qwen35}. Models decode greedily through vLLM \citep{kwon2023vllm}, with one explanation-first template per LLM shared across methods. Prompts are capped at 32,256 tokens. Output caps are 512 tokens for Qwen2.5-7B and 2,048 for the other LLMs, held constant within each comparison. Main indexes use seed 0; seeds 1 and 2 are controls. 


\begin{table*}[t]
\centering
\footnotesize
\setlength{\tabcolsep}{5pt}
\begin{tabular}{@{}l rrr rrr rrr@{}}
\toprule
& \multicolumn{3}{c}{TQA-Bench (1,200)} & \multicolumn{3}{c}{BirdQA-R (304)} & \multicolumn{3}{c}{DataBench-Test-L (311)} \\
\cmidrule(lr){2-4}\cmidrule(lr){5-7}\cmidrule(l){8-10}
Method & Q2.5 & L3.1 & Q3.5 & Q2.5 & L3.1 & Q3.5 & Q2.5 & L3.1 & Q3.5 \\
\midrule
Truncated table & 66.2 & \underline{71.0} & 84.1 & 17.4 & 13.5 & 23.4 & 31.2 & \underline{28.0} & 23.5 \\
BM25-Nat & 70.6 & 67.5 & \underline{86.5} & \underline{18.1} & 16.8 & \underline{25.3} & \underline{31.5} & 22.5 & 24.4 \\
BM25-2K & \underline{71.0} & 64.9 & 80.2 & 16.8 & \underline{17.4} & \underline{25.3} & 26.4 & 23.8 & \underline{28.9} \\
Dense-2K & 54.3 & 51.4 & 59.7 & 13.8 & 13.5 & 20.1 & 25.4 & 19.0 & 23.2 \\
LLMLingua-2 & 57.0 & 51.5 & 61.7 & 17.4 & 11.5 & 14.1 & 17.7 & 13.8 & 12.9 \\
TAP4LLM & 49.8 & 50.4 & 53.8 & 15.5 & 14.1 & 17.1 & 24.8 & 21.9 & 23.2 \\
\midrule
TAP4LLM+Stats (2K)$^\ddagger$ & 47.5 & 52.8 & 58.1 & 16.4 & 14.8 & 19.7 & 35.4 & 31.8 & 29.9 \\
\midrule
\TR{} & \textbf{75.9}$^\dagger$ & \textbf{76.6}$^\dagger$ & \textbf{87.8} & \textbf{34.5}$^\dagger$ & \textbf{30.9}$^\dagger$ & \textbf{41.8}$^\dagger$ & \textbf{42.4}$^\dagger$ & \textbf{36.7}$^\dagger$ & \textbf{47.3}$^\dagger$ \\
\bottomrule
\end{tabular}
\caption{Observed accuracy (\%) across three LLMs and benchmarks. Q2.5, L3.1, and Q3.5 denote Qwen2.5-7B, Llama-3.1-8B, and Qwen3.5-9B. Underlines mark the strongest of six primary baselines. $\dagger$: paired question-level intervals against all six baselines exclude zero. $\ddagger$: strengthened TAP4LLM control, outside the 54-comparison family (Table~\ref{tab:tapstats}). Internal configurations appear in Table~\ref{tab:internal}.}
\label{tab:main}
\end{table*}


\paragraph{Comparators.} The six external baselines are truncated-table prompting, BM25-Nat and BM25-2K \citep{robertson2009bm25}, Dense-2K \citep{karpukhin2020dpr,xiao2024cpack}, LLMLingua-2 \citep{pan2024llmlingua2}, and TAP4LLM \citep{sui2024tap4llm}. Truncation retains a roughly 24K-token prefix; BM25-Nat uses a question-dependent budget capped at 22K. BM25-2K and Dense-2K pack ranked rows within 2K tokens. LLMLingua-2 compresses BM25-Nat; main-table TAP4LLM uses similarity sampling, size/type augmentation, and 1K packing. A strengthened TAP4LLM control adds full column statistics and 2K packing. Internal policies reuse \TR{}'s components to compare representation and budget (Appendix~\ref{app:internal}).

\paragraph{Scoring and uncertainty.} TQA-Bench uses the first standalone answer-option letter. BIRD and DataBench use normalized denotation accuracy with multiset comparison and numerical tolerance. We report paired 95\% bootstrap intervals over questions with 1,000 resamples \citep{efron1993bootstrap}, a separate table-level bootstrap, and Holm correction for the 54 main baseline comparisons (10,000 resamples). Differences are computed before rounding. Component and diagnostic comparisons are outside that 54-comparison family (Appendix~\ref{app:scoring}).

\paragraph{Compute.} For table $T$ serving $Q_T$ questions, estimated per-question cost is
\begin{equation}
\begin{aligned}
\mathcal{F}(q,T) ={} & \frac{\mathcal{F}_{\mathrm{index}}(T)}{Q_T} + \mathcal{F}_{\mathrm{online}}(q,T)\\
& + \mathcal{F}_{\mathrm{LLM}}(q,T).
\end{aligned}
\label{eq:accounting}
\end{equation}
LLM cost includes prefill and decoding; preprocessing includes embeddings and clustering where used. We average per-question estimates, not the cost at the mean prompt length. Appendix~\ref{app:compute} details FLOP accounting,
omitted CPU lexical-index costs, and warm-/cold-start
timing.

\section{Results}
\label{sec:results}

\subsection{Accuracy across Benchmarks and LLMs}
\label{sec:perf}

\TR{} achieves \textbf{higher observed accuracy than the strongest of six primary single-call baselines in all nine settings}, by 1.2--18.3 accuracy points (Table~\ref{tab:main}). With Qwen2.5-7B, accuracies are 75.9\%, 34.5\%, and 42.4\% on TQA-Bench, BirdQA-R, and DataBench-Test-L, using mean prompts of 1,579, 941, and 1,066 tokens, compared with
approximately 20--24K tokens for truncated-table prompting. Of 54 paired comparisons, 53 remain significant after Holm correction; 52 also have table-level intervals excluding zero. The sole inconclusive Holm comparison is Qwen3.5-9B on TQA-Bench: \TR{} achieves 87.8\% versus 86.5\% for BM25-Nat, with 1,601 versus 6,475 mean prompt tokens. Both table-level exceptions concern Qwen3.5-9B on TQA-Bench (truncation and BM25-Nat; Appendix~\ref{app:scoring}).

On BirdQA-R and DataBench-Test-L, even the least accurate \TR{} LLM exceeds the strongest primary baseline evaluated with any of the three LLMs, suggesting that its advantage is not merely attributable to the choice of LLM. Adding full-column statistics and a 2K budget strengthens TAP4LLM in eight of nine settings; nevertheless, \TR{} leads the strengthened control by 4.8--29.7 percentage points (Table~\ref{tab:tapstats}). Furthermore, BirdQA-R provides additional evidence hints for several questions, identically across methods. To assess whether our gains depend on these hints, we rerun \TR{} and all six primary baselines without them. When hints are removed \emph{from every method} on Qwen2.5-7B \TR{} still scores 29.3\%, versus 14.8--18.8\% for the six re-run primary baselines (Appendix~\ref{app:robust}).

\paragraph{Scale and prompt robustness.} On the descriptive subset of DataBench-Test-L tables exceeding 100,000 rows, \TR{} achieves 51.4\%, 47.2\%, and 61.1\% accuracy across three LLMs, versus 33.3\%, 26.4\%, and 29.2\% for their strongest primary baselines, providing evidence of \emph{scale robustness} (Appendix~\ref{app:size}).

To examine whether long explanations and output-token limits contribute to the observed gains, we repeat the comparison using an answer-only prompt with Qwen3.5-9B. Under this setting, \TR{} achieves 37.2\% on BirdQA-R and 50.8\% on DataBench-Test-L, outperforming the strongest of tested baselines by 19.1 and 12.9 points, respectively. These results show that the advantage persists without explanation-first prompting (Appendix~\ref{app:robust}).

\begin{figure*}[t]
\centering
\includegraphics[width=\textwidth]{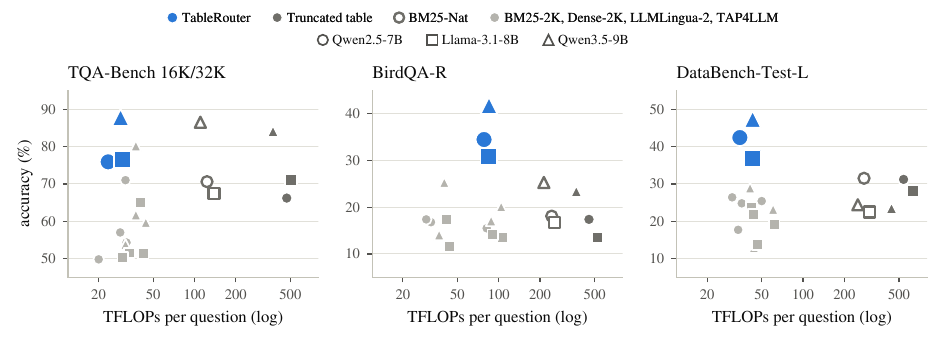}
\caption{Accuracy versus estimated compute for the main external-baseline comparison with Qwen2.5-7B, Llama-3.1-8B, and Qwen3.5-9B (logarithmic horizontal axis). Costs include LLM prefill/decoding and preprocessing as defined in Eq.~\eqref{eq:accounting}, with table indexes amortized across questions. Internal configurations are compared in Table~\ref{tab:internal} and Appendix~\ref{app:visual}.}
\label{fig:pareto}
\end{figure*}

\subsection{Post-Freeze Evaluation on Long Tables}
\label{sec:fresh}

To assess whether the gains extend beyond development-informed benchmarks, we evaluate \TR{} on post-freeze BIRD. With Qwen2.5-7B, \TR{} answers 107 questions correctly (35.8\%), compared with 49 (16.4\%) for LLMLingua-2, the strongest external baseline (Table~\ref{tab:fresh}). The +19.4-point gain is supported by positive 95\% bootstrap intervals at both the question ([13.7, 25.8]) and table ([9.2, 29.2]) levels. All six external comparisons remain significant after a separate Holm correction.

The SQL-derived nature of post-freeze BIRD makes full-table computation particularly relevant. Consistent with this, \TR{} selects the computation route for 206 of 299 questions (68.9\%) and achieves 35.8\% accuracy. This substantially exceeds Coverage Alone (20.1\%) and is close to the 35.5\% achieved by the more permissive Computation-First policy (Table~\ref{tab:fresh}). These results highlight the importance of computed evidence on database-grounded questions, while \TR{} retains the flexibility to select coverage-based evidence for other queries.

\begin{table}[t]
\centering
\footnotesize
\setlength{\tabcolsep}{4pt}
\begin{tabular}{@{}lrr@{}}
\toprule
Method & Accuracy & Prompt tokens \\
\midrule
Truncated table & 9.0 & 23,479 \\
BM25-Nat & 12.7 & 12,399 \\
BM25-2K & 13.4 & 2,144 \\
Dense-2K & 11.7 & 2,142 \\
LLMLingua-2 & 16.4 & 1,517 \\
TAP4LLM & 13.0 & 1,202 \\
\midrule
Coverage Alone & 20.1 & 1,931 \\
Computation-First & 35.5 & 654 \\
Hybrid-2000 & 32.8 & 1,953 \\
\midrule
\TR{} & \textbf{35.8} & 969 \\
\bottomrule
\end{tabular}
\caption{Post-freeze BIRD results with Qwen2.5-7B:
accuracy (\%) and mean prompt tokens across
299 questions and 55 tables.}
\label{tab:fresh}
\end{table}

\subsection{Compute and the Repeated-Query Workload}
\label{sec:compute}

With table-index construction amortized across questions, \TR{} requires \textbf{77--95\% fewer estimated FLOPs} than truncated-table prompting and 60--88\% fewer than
BM25-Nat across all nine settings (Figure~\ref{fig:pareto}). These savings extend to
post-freeze BIRD, where \TR{} requires 85.2\% and 68.5\% fewer estimated FLOPs than truncation and BM25-Nat, respectively. \TR{} also achieves higher accuracy with fewer estimated FLOPs than Dense-2K in all nine settings. Some lighter baselines remain cheaper, but achieve lower accuracy.

For repeatedly queried tables, a reusable index also reduces online latency. On an NVIDIA A100 80GB, median Qwen2.5-7B warm-index latencies are 0.8, 0.9, and 0.4 seconds on TQA-Bench, BirdQA-R, and DataBench-Test-L, respectively, compared with 3.1, 1.5, and 0.9 seconds for truncation. These latency results assume a prebuilt index and do not establish a cold-start advantage (Appendix~\ref{app:compute}).

\section{What Makes the Evidence Useful?}
\label{sec:analysis}

\subsection{Computed Statistics Improve Evidence Use}
\label{sec:computed}

The fixed-row experiment reveals a gap between \emph{evidence access} and \emph{evidence use}. When the same answer-determining rows are provided, adding their computed statistics \textbf{doubles Qwen2.5-7B accuracy from 40.0\% to 80.0\% (+40.0 points)} and improves Llama-3.1-8B \textbf{from 37.8\% to 71.1\% (+33.3 points)} (Table~\ref{tab:fixedrows}). On these 45 questions, the gains correspond to 18 and 15 additional correct answers, respectively. Qwen3.5-9B already achieves 86.7\% with rows alone and improves to 93.3\% with computed statistics.

These results show that providing all the necessary rows does not guarantee that an LLM will use them successfully. Because the rows are identical, the improvements cannot be attributed to retrieving additional evidence. Instead, making the required computations explicit can substantially improve the LLM's ability to answer from the available data. The smaller observed gain for Qwen3.5-9B is consistent with its stronger performance on rows alone.

The control uses 45 gold-query-selected questions (41 counts and four sums), and the added statistics may directly provide the requested answer. It therefore demonstrates the value of explicit computation, not automatic condition extraction or unrestricted numerical reasoning. Complementary comparisons under a shared 2K-token budget also find 6.2--14.1-point accuracy gains from combining computed statistics with coverage evidence across all three LLMs on BirdQA-R and DataBench-Test-L (Appendix~\ref{app:internal}).

\begin{table}[t]
\centering
\footnotesize
\setlength{\tabcolsep}{4pt}
\begin{tabular}{@{}lrrl@{}}
\toprule
LLM & Rows & +Statistics & Gain [95\% CI] \\
\midrule
Qwen2.5-7B & 40.0 & 80.0 & $+40.0$ [26.7, 55.6] \\
Llama-3.1-8B & 37.8 & 71.1 & $+33.3$ [17.8, 48.9] \\
Qwen3.5-9B & 86.7 & 93.3 & $+6.7$ [0.0, 15.6] \\
\bottomrule
\end{tabular}
\caption{Fixed-row evidence-use control on 45
BirdQA-R questions: accuracy (\%) with identical
answer-determining rows, without and with computed
statistics. Intervals are paired 95\% question-level
bootstrap intervals. Selection uses gold queries;
contexts are not cost-matched.}
\label{tab:fixedrows}
\end{table}

\subsection{Different Questions Favor Different Evidence}
\label{sec:policies}

The benefits of computed statistics depend on the question. Lookup questions often require original row-level associations, whereas aggregation questions can benefit from explicitly computed values. We examine this trade-off by comparing \TR{} with Coverage Alone, which retains retrieved rows and column summaries, and Computation-First, which selects computation contexts more readily than \TR{} and otherwise falls back to coverage.

The results reveal distinct strengths (Table~\ref{tab:internal}). On lookup-oriented TQA-Bench, Coverage Alone achieves 79.8\% accuracy, compared with 64.0\% for Computation-First. This suggests that preserving original rows is particularly valuable when answers depend on identifying specific records. The pattern changes on BirdQA-R, where Computation-First achieves 35.5\%, compared with 24.7\% for Coverage Alone. On DataBench-Test-L, \TR{} achieves the highest accuracy at 42.4\%, versus 40.8\% for Computation-First and 32.5\% for Coverage Alone. These results highlight the value of computed evidence on database-grounded questions, while also showing that the preferred representation varies across benchmarks. 

The strength of \TR{} lies in its ability to handle diverse question types with a single query-aware policy. On lookup-oriented TQA-Bench, it outperforms Computation-First by 11.9 accuracy points, while on BirdQA-R and DataBench-Test-L, it exceeds Coverage Alone by 9.8 and 9.9 points, respectively. Across all three benchmarks, \TR{} also requires less estimated compute than Coverage Alone. Together, these results demonstrate that \TR{} provides a \textbf{strong and consistent accuracy--compute balance} across heterogeneous table-question answering tasks, without requiring dataset-specific policy selection. Further comparisons appear in Appendix~\ref{app:internal}.

\begin{table}[t]
\centering
\footnotesize
\setlength{\tabcolsep}{3.5pt}
\begin{tabular}{@{}lccc@{}}
\toprule
Configuration & TQA & Bird & DB-L \\
\midrule
\TR{} & 75.9/23.5 & 34.5/78.1 & 42.4/34.6 \\
Coverage Alone & 79.8/27.7 & 24.7/90.5 & 32.5/48.1 \\
Computation-First & 64.0/8.4 & 35.5/73.6 & 40.8/29.7 \\
\bottomrule
\end{tabular}
\caption{Evidence-construction policies with
Qwen2.5-7B: accuracy (\%) / estimated TFLOPs
per question.}
\label{tab:internal}
\end{table}

\section{Conclusion}
\label{sec:conclusion}

Long-table question answering requires more than retrieving relevant rows: incomplete evidence can prevent exact computation, while even complete evidence may be difficult for an LLM to use. We introduced \TR{}, a training-free framework that constructs query-dependent evidence by selecting between full-table computed statistics and coverage-based retrieval, using a single frozen-LLM call.

Across three LLMs and three benchmarks, \TR{} outperforms six baselines in all nine settings, requiring 77--95\% fewer estimated FLOPs than truncated-table prompting with amortized indexing. The gains extend to post-freeze long-table evaluation. Controlled experiments further show that computed statistics substantially improve accuracy, even when all answer-determining rows are available. These findings establish evidence construction as a central challenge in long-table QA: effective reasoning depends not only on which rows are retrieved, but on \emph{what is computed, what is preserved, and what is made explicit to the LLM}.

\section*{Limitations}

The study covers English benchmarks, three open-weight LLMs of 7--9B parameters, and a single-call interface. The post-freeze BIRD evaluation uses Qwen2.5-7B alone. TQA-Bench and BirdQA-R were development-informed; DataBench test inputs and some labels underwent LLM-free processing. We therefore report a separate post-freeze BIRD evaluation on databases not used during development; this evaluation uses Qwen2.5-7B alone and cannot rule out exposure during model pretraining. DataBench-Test-L contains nine tables, limiting fine-grained generalization across table populations (Appendix~\ref{app:protocol}).

Query matching and routing rely on lexical cues, which may miss paraphrases or misinterpret filtering conditions. The route-specific token budgets were chosen heuristically, and alternative budgets may affect the accuracy--compute trade-off.
The fixed-row experiment demonstrates the benefit of explicit computations with identical answer-determining rows, but uses 45 gold-query-selected questions, predominantly counts; the added statistics may directly provide the answer, and the contexts are not compute-matched. Shared-budget comparisons may also change retained rows, while component ablations do not establish independent gains from every heuristic (Appendices~\ref{app:internal} and~\ref{app:diag}). The complete system is evaluated as implemented; a retrospectively simplified system is not substituted for it. The current experiments also do not isolate every possible combination of simple retrieval and the same fact constructor. 

Estimated FLOPs include amortized indexing but
omit some CPU lexical-index costs and do not
measure energy consumption. Warm-index latency
captures the benefits of repeated-query serving,
whereas index construction adds cold-start cost.
Timing measurements use sampled questions with
one LLM, and bootstrap intervals do not capture
all serving variability. Program-executing and multi-call systems are outside the evaluated interface. (Appendix~\ref{app:compute}).

\section*{Ethical Considerations}

Our experiments use publicly released benchmarks under their stated licenses, with personal details redacted from audit materials. Nevertheless, both retrieved records and statistics computed over narrowly defined groups may reveal sensitive information. Responsible deployment should therefore preserve the access permissions of the underlying data, minimize unnecessary disclosure, and apply appropriate safeguards to small-group statistics. 
\bibliography{references}

\clearpage
\appendix

\section{Evaluation Populations and Protocol}
\label{app:protocol}

The main comparison evaluates the same primary \TR{} policy for all
three LLMs and benchmarks. Its computation and coverage contexts have
600- and 2,000-token evidence caps, respectively. We distinguish the
main development-informed evaluations from the separately committed
post-freeze BIRD population.

\subsection{Dataset Construction}

\paragraph{TQA-Bench.}
The main population comprises 1,200 first-template entity-lookup
questions on 120 tables from the 16K/32K settings. Related tables are
stacked identically for all methods. Development additionally uses
600 questions on 60 tables from the 64K setting. The main tables
contain fewer than 10,000 rows; their difficulty is characterized
by serialized length rather than a minimum row count.

\paragraph{BirdQA-R.}
Starting from 361 single-table BIRD questions, we append available
evidence hints, recompute gold answers on supplied tables capped at
200,000 rows, preserve complete list answers, and exclude answers
exceeding 192 tokens. This leaves 317 questions; excluding 13
previously inspected during development yields \textbf{304 questions
on 52 tables}. Of these, 272 have hints, supplied identically to all
methods. The tables range from 25 to 200,000 rows; 99 questions on
19 tables involve fewer than 1,000 rows.

\paragraph{DataBench.}
The test split has 522 questions on 15 tables. A predeclared cutoff
of 32,576 Qwen2.5-7B tokens for full Markdown serialization selects
nine tables and 311 questions for DataBench-Test-L; the final
32,256-token prompt limit yields the same partition. All selected
tables are used in full, spanning 517--239,461 rows and 5--35 columns.
The remaining six tables form Test-S (211 questions), and the
development split contains 320 questions. One development table,
\texttt{065\_RFM}, was capped at 200,000 of its 541,909 rows despite
full-table gold answers for 20 questions; this affects development
analysis, not the Test-L evaluation.

\paragraph{Post-freeze BIRD.}
The protocol starts from the BIRD training release's 69 databases,
excluding ten used by earlier explorations and six supplying TQA-Bench.
Within the remaining 53 databases, 354 single-table queries reference
tables with at least 10,000 and fewer than 200,000 rows. Excluding
8 empty answers, 3 answers sensitive to missing-cell interpretation,
and 44 answers longer than 192 tokens gives \textbf{299 questions on
55 tables from 27 databases}. Their row counts range from 10,000 to
186,697; 288 questions have hints. The question population was
committed after the method freeze and before indexing or LLM calls,
and the six primary external comparisons were specified before the
evaluation. A stored-run audit found no earlier evaluations from these
databases on the authors' machines. This is evidence about recorded
development exposure, not about the LLMs' pretraining data.

\subsection{Development Exposure and Evaluation Status}

Earlier explorations evaluated the main TQA-Bench and BirdQA-R
questions before the rules were frozen on 5 October 2026. Grounding,
route selection, and two implementation corrections were informed
by these earlier analyses; the main benchmark matrix is therefore
\emph{development-informed}. No LLM answered DataBench test questions
before the freeze, although LLM-free processing inspected test inputs
and some labels. In a sensitivity analysis excluding the 86 Test-L
questions with prior gold-string checks, \TR{} retains positive paired
question-level intervals against all six external baselines on the
remaining 225 questions. The strengthened TAP4LLM comparison on that
subset is inconclusive: $+6.7$ [$-0.9$, $+13.8$] points.

A separate post hoc TQA-Bench check removes 70 wording/option matches
to an earlier third-party dataset copy. Across the three LLMs, the
remaining 1,130 questions show observed leads of 5.4, 7.8, and
1.3 accuracy points over the strongest external baseline; 17 of 18
paired intervals exclude zero. These checks characterize sensitivity
to documented exposure rather than retroactively creating untouched
test sets. The Qwen2.5-7B shared-budget comparison was post hoc;
extensions to the two other LLMs were registered before their context
construction, but after the Qwen2.5-7B result. Prompt and routing
controls introduced later are reported as supplementary analyses.

\section{Implementation and Evidence Construction}
\label{app:impl}

The description below specifies the implemented primary configuration
without changing the route or parameters used for the main results.
All evidence caps exclude the shared question and answering instructions,
which are included in reported prompt lengths. The LLM is frozen and
called once per question.

\subsection{Online Procedure and Index}

\begin{algorithm}[t]
\caption{\TR{}: query-dependent evidence construction}
\label{alg:primary}
\small
\begin{algorithmic}[1]
\Require Question $q$, supplied table $T$, reusable index $\mathcal I_T$
\State $\bar q\gets q$ without answer options
\State Build question vector $u$, anchors, candidates $\mathcal C$, and scores $p_i$
\State Extract conditions, operations, and target columns
\State $\Phi\gets$ first 18 statistic lines computed over $T$
\State $G\gets$ lines passing lexical grounding
\State $b\gets (n\le60)$ and full table fits 600 tokens
\If{$\neg b\land\mathrm{comp}(\bar q)\land |G|>0$}
  \State $X\gets$ grounded statistics and optional supporting rows (600 tokens)
\ElsIf{full table fits in 2,000 tokens}
  \State $X\gets$ full table serialization
\Else
  \State Form coverage needs $\mathcal J$ and candidate units $\mathcal E$
  \State Greedily add affordable units by positive gain per token
  \State Fill remaining space with ranked rows; restore table order
  \State $X\gets$ selected summaries followed by rows (2,000 tokens)
\EndIf
\State \Return answer from the frozen LLM given $(q,X)$
\end{algorithmic}
\end{algorithm}

Each row is serialized with column names and type-normalized values,
using \texttt{[COL]}, \texttt{[VAL]}, and \texttt{[SEP]} delimiters.
BGE-small encodes at most 512 input tokens per row, producing unit
embeddings $x_i$. Spherical $k$-means builds primary and secondary
cluster assignments and normalized prototype vectors. The initial and
minimum prototype counts are
\begin{equation}
\begin{aligned}
K_0 &= \min\{n,\max(4,\lfloor\sqrt n\rfloor)\},\\
K_{\min} &= \max\!\left\{1,\min\!\left(K_0,
 \max\!\left(4,\left\lfloor K_0/4\right\rceil\right)\right)\right\},
\end{aligned}
\label{eq:kmin}
\end{equation}
where $\lfloor\cdot\rceil$ denotes nearest-integer rounding, with
ties to even. Initialization uses seed 0 and at most 25 spherical
$k$-means iterations. Up to six split rounds require a parent of
at least 16 rows, two children of at least eight rows, a weighted
cohesion gain of at least 0.05, and no more than 256 prototypes after
the split. Closest prototypes merge while cohesion loss is at most
0.02, stopping at $K_{\min}$ or a failed merge. Final assignments
limit cluster size to $\lceil1.5n/K\rceil$; all reported tables settle
at $K_{\min}$. The 256 restriction applies to accepted splits,
not necessarily to the initial $K_0$.

Prototypes are ranked by Eq.~\eqref{eq:prototype_weights} until their
weights sum to at least 0.95. Rows belonging to selected primary or
secondary clusters are retrieval candidates. On tables with $n\ge500$,
we intersect this set with the best $\min(2000,\lfloor n/2\rfloor)$
BM25 rows, reverting to prototype candidates if the intersection is
empty. Rare-cell and selective-anchor matches of at most 50 rows are
added. Aggregation-cue questions and tables with $n<30$ use all rows
as retrieval candidates; this is distinct from the small-table
\emph{routing} bypass.

\begin{table}[t]
\centering
\footnotesize
\setlength{\tabcolsep}{2.5pt}
\resizebox{\columnwidth}{!}{%
\begin{tabular}{@{}lr@{}}
\toprule
Parameter & Value \\
\midrule
Computation / coverage evidence caps & 600 / 2,000 tokens \\
Statistic lines / conditions / hops & 18 / 4 / 2 \\
Numeric / categorical targets & 3 / 2 \\
Encoder input cap & 512 tokens \\
Small-table computation bypass & $n\le60$, fits 600 \\
Selective anchor & $|R_t|\le\max(5,0.2n)$ \\
Anchor weight range & [0.2, 1] \\
Prototype temperature / size penalty & 0.1 / 0.05 \\
Cumulative prototype mass & 0.95 \\
Row-score weights & 1 / 0.5 / 1 \\
Row-ranking temperature & 0.1 \\
Non-agg. row threshold & $p_i\ge0.35\max p$ \\
Minimum / maximum kept rows & 8 / 500 \\
Aggregation posterior mass & 0.999 \\
All-row retrieval exception & $n<30$ \\
BM25 candidate cutoff ($n\ge500$) & $\min(2000,\lfloor n/2\rfloor)$ \\
BM25 $k_1$ / $b$ & 1.5 / 0.75 \\
Column mass / temperature & 0.96 / 0.25 \\
Cluster split gain / merge loss & 0.05 / 0.02 \\
\bottomrule
\end{tabular}}
\caption{Implementation constants for the primary \TR{} policy.}
\label{tab:constants}
\end{table}

\subsection{Question Anchors and Coverage Packing}

Question anchors are literal spans matching table values: quoted
phrases, numbers, dates, capitalized entities, or up to five-word
phrases, excluding column names. A matching value $t$ is selective
when $|R_t|\le\max(5,0.2n)$. Its rarity weight and the normalized
row-level anchor score are
\begin{equation}
\begin{aligned}
w_t &= \min\!\left(1,\max\!\left(0.2,
 \frac{\ln(n/|R_t|)}{\ln\max(n,2)}\right)\right),\\
\eta_i &= \frac{\sum_{t:i\in R_t}w_t}{\sum_t w_t}.
\end{aligned}
\label{eq:anchor}
\end{equation}
The latter is defined for a nonempty set of extracted selective
anchors. The $p_i$ values in Eq.~\eqref{eq:posterior} are
\emph{ranking weights}, not probabilities of answer correctness.

Column ranking combines header embeddings, lexical relevance, a
type prior, and value matches. At most eight-column tables retain
all columns. Larger tables retain columns accounting for mass 0.96
at temperature 0.25, plus named columns and an identifier; if fewer
than three remain, the first eight are added. Column scores also
assist target selection independently of the row distribution.

Coverage units comprise candidate rows, numeric-column summaries,
and a block containing all summaries; the summaries describe entire
columns, not question-conditioned subsets. The objective in
Eq.~\eqref{eq:objective} rewards complementary matches to selective
anchors and requested operations. A greedy heuristic selects units
with positive marginal coverage per token. Non-aggregation fill uses
rows with $p_i\ge0.35\max p$, with a minimum pool of eight where
available; aggregation fill retains the smallest ranked set of mass
at least 0.999. Two neighbors per selected row can be added, capped
at 500 candidate rows before final packing. Rows that do not fit are
skipped; surviving rows are restored to source order. The eight-row
pool threshold does not guarantee eight rows in the packed context,
and the employed heuristic carries no claimed approximation factor
\citep{khuller1999budgeted}.

\subsection{Full-Table Statistics, Grounding, and Packing}

The constructor considers at most four extracted conditions,
three numerical targets, two categorical targets, and two identifier
link hops. It computes over the \emph{matching rows of the full supplied
table}, not the retrieval pool. Candidate lines may include counts,
sums, means, extrema, medians, frequency counts, conditional shares,
ratios, differences, or correlations, depending on cues and available
types. Numerical aggregates operate on finite target values and
retain repeated values. A line may contain several statistics;
at most 18 candidate lines enter grounding.

Lexical grounding checks the question text without options for named
columns, simple inflections, supported header expansions, and counting
patterns. Conditioned counts are permitted for count, ratio, and yes/no
cues when an extracted filter is available. A correlation requires
both target columns to be mentioned. These checks establish a lexical
connection to the question, not that every computed value correctly
answers it. Table~\ref{tab:cues} records the principal implementation
patterns.

\begin{table*}[t]
\centering
\footnotesize
\setlength{\tabcolsep}{4pt}
\renewcommand{\arraystretch}{1.08}
\begin{tabular}{@{}p{0.15\textwidth}p{0.79\textwidth}@{}}
\toprule
Cue family & Implementation words and patterns \\
\midrule
Aggregation & sum, total, combined, overall, average, avg, mean, median, count, how many, number of, difference \\
Extremes & highest, largest, maximum, most, biggest, greatest, top, best, lowest, smallest, minimum, least, fewest, worst; top, first, bottom or last $k$; $k$ highest; highest $k$ \\
Count & how many, number of, count \\
Frequency & most common, frequent, popular or often; mode; appears most; occurs most; most \dots{} entries, records or rows; least or fewest \dots{} common, frequent, popular or often \\
Ratio & ratio, percent, percentage, proportion, share, fraction, times as, times more, times less \\
Difference & difference, differ, subtract\dots, minus, how much or many more, less, fewer, higher or lower \\
Correlation & correlat\dots, relationship between \\
Comparison & more, greater, higher, larger, bigger, older, longer, taller, heavier, above, over or exceeds (than) $x$; at least, no less than or minimum of $x$; less, fewer, lower, smaller, younger, shorter, lighter, cheaper, below or under (than) $x$; at most, no more than or maximum of $x$; $x$ or more, or less; $\ge,\le,>,<$ $x$; between $x$ and $y$ \\
Yes/no & question begins with is, are, was, were, does, do, did, has, have, had, can or will \\
Scope & all, every, entire, whole, each; coverage route only \\
\bottomrule
\end{tabular}
\caption{Lexical cue families. Whole-word matching is used except for the prefix stems \texttt{subtract} and \texttt{correlat}; comparison symbols are recognized directly.}
\label{tab:cues}
\end{table*}

Computation contexts prioritize conditioned statistic lines and may
include original rows supporting extrema, links, or anchors. Packing
first removes supporting rows, then ungrounded lines, and finally
lower-priority grounded lines that exceed the cap. Every observed
computation context retains at least one grounded line. Not all
candidate grounded lines fit in 37/244, 48/200, and 11/204
computation-assigned questions on TQA-Bench, BirdQA-R, and
DataBench-Test-L, respectively. These counts describe packing,
not the counterfactual accuracy effect of omitted lines.

\section{Baselines, Scoring, and Statistical Tests}
\label{app:scoring}

\subsection{Shared Inputs and Evaluation}

All evaluated systems receive the same supplied table, question,
available hint, and per-LLM answer template. They need not receive
the same \emph{LLM-visible evidence}: ordering, length, retrieved
cells, and explicit computation differ. Truncation keeps the header
and a source-order prefix of complete rows under its approximately
24K-token rule. BM25-2K and Dense-2K pack lexically and semantically
ranked rows, respectively, within 2K tokens. BM25-Nat uses a
question-dependent cap of at most 22K with complete retained columns.
LLMLingua-2 compresses its input toward 2K; primary TAP4LLM uses
similarity sampling, size/type augmentation, and 1K packing.
The strengthened TAP4LLM statistics control is evaluated separately.
BM25-Nat is not charged the \TR{} index construction.

BIRD and DataBench scores use normalized answer denotations as
\emph{multisets}, preserving duplicates but ignoring item order.
Strings are normalized for case, quotes, brackets, and formatting;
yes/no values map to true/false. Numeric comparison has relative
tolerance $10^{-6}$ and accepts a gold value that rounds to a
prediction displayed with at least two decimals. TQA-Bench extracts
the first standalone multiple-choice letter. An output without an
extractable final answer is incorrect.

\subsection{Uncertainty, Correction, and Repeated Serving}

Paired 95\% bootstrap intervals resample individual questions 1,000
times while retaining paired outcomes. A complementary bootstrap
resamples whole tables with their associated questions, reflecting
between-table variation. The main $3\times3\times6=54$ external
comparisons receive Holm correction using bootstrap $p$-values from
10,000 resamples. Post-freeze BIRD uses a separate six-comparison
family; exploratory policy and component comparisons do not enter
either correction family. Differences are computed from unrounded
accuracies with ties-to-even final rounding, so subtraction of
one-decimal cells can differ from the paired difference by 0.1 point.

Of 54 main external comparisons, \textbf{53 remain significant after
Holm correction}. The exception is Qwen3.5-9B against BM25-Nat on
TQA-Bench: $+1.2$ [$-1.3$,$+3.8$] points, uncorrected $p=0.30$.
Table-level intervals exclude zero in 52 comparisons; exceptions
are Qwen3.5-9B against truncation [$-1.4$,$+8.5$] and BM25-Nat
[$-4.2$,$+6.7$] on TQA-Bench. Zero-spanning intervals are
inconclusive, not equivalence tests. All six post-freeze external
comparisons pass their separately applied Holm correction.

Changed prompts are rerun in the component controls; byte-identical
prompts reuse stored answers. An auxiliary reconstruction check
verified prompt lengths rather than bytewise context identity.
Repeated serving of 50 Llama-3.1 prompts gave 64--66\% accuracy.
Consequently, bootstrap intervals describe sampling uncertainty,
not every source of generation or serving variability.

\section{Controlled Evidence Comparisons}
\label{app:internal}
\label{app:visual}

These controls separate two questions: whether an explicit computation
helps \emph{after evidence access is fixed}, and whether computed
statistics add value to retrieval within a fixed token budget.
Additional policy comparisons examine why neither representation is
appropriate for every question. Each figure is based on the reported
runs; none introduces additional LLM evaluations.

\subsection{When the Necessary Rows Are Already Available}

The fixed-row control selects 45 BirdQA-R single-aggregation questions
using gold queries, independently of \TR{} output. All qualifying
rows, including named query columns and an identifier, fit in 2K
tokens. In the Rows condition, the LLM receives the same
answer-determining rows without the computed result; in +Statistics,
count, sum, mean, minimum, and maximum are additionally supplied.
The 41 counts and four sums make the correct aggregate potentially
explicit, but the original rows remain identical.

\begin{figure*}[t]
\centering
\includegraphics[width=0.94\textwidth]{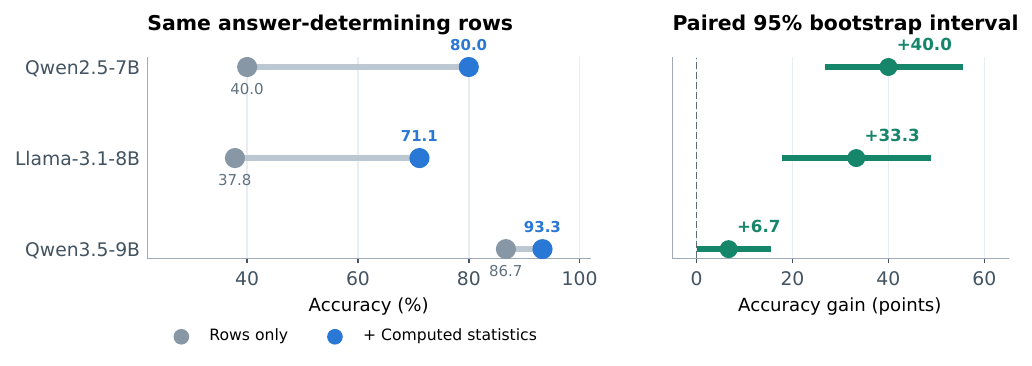}
\caption{Evidence use with identical rows. Left: rows-only versus rows plus computed statistics. Right: paired accuracy gains and 95\% question-level bootstrap intervals on 45 selected BirdQA-R questions.}
\label{fig:j05}
\end{figure*}

\textbf{Making the computation explicit increases accuracy without
retrieving more rows.} Figure~\ref{fig:j05} shows that Qwen2.5-7B
gains 18 correct answers and Llama-3.1-8B gains 15 across the 45
questions. Their paired 95\% intervals exclude zero. Qwen3.5-9B
begins with substantially higher rows-only accuracy; its observed
$+6.7$-point gain has an interval that reaches zero. The manipulation
is not compute matched, and it tests the usefulness of an explicit
calculation rather than automatic semantic parsing.

\subsection{Computed Statistics within a Shared Evidence Budget}

To test whether this benefit survives when statistics must share
space with retrieval, a 2,000-token combined control places available
computed lines before coverage evidence. We call this control
\emph{Hybrid-2000}; it is a single combined-context comparison, not
a replacement for the primary \TR{} gate. Coverage Alone uses the
same cap without the question-conditioned lines. When no lines are
available, their contexts coincide and stored answers are reused.

\begin{figure*}[t]
\centering
\includegraphics[width=0.94\textwidth]{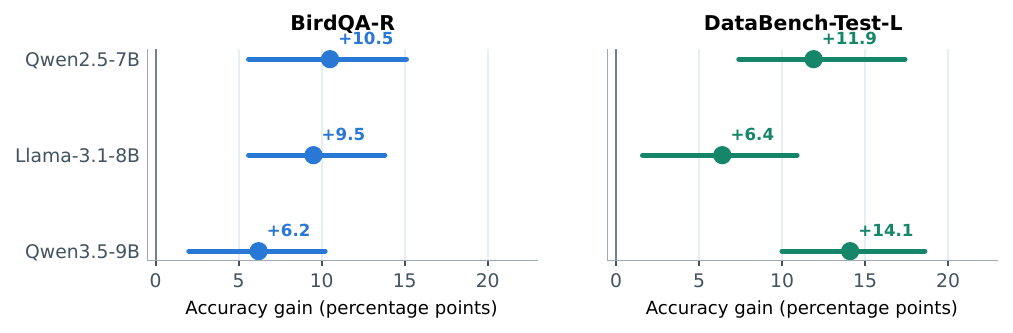}
\caption{Hybrid-2000 minus Coverage Alone at a shared 2K evidence cap. Each point shows the accuracy gain, with its paired 95\% question-level interval. Both benchmarks have positive gains for all three LLMs.}
\label{fig:j06}
\end{figure*}

\textbf{Computed evidence improves both database-grounded benchmarks
across all three LLMs.} On BirdQA-R and DataBench-Test-L,
Hybrid-2000 improves accuracy over Coverage Alone by 6.2--14.1
points (Figure~\ref{fig:j06}, Table~\ref{tab:hybcov}). All six
question-level and six table-level 95\% intervals exclude zero.
Unlike the fixed-row control, contexts can retain different rows:
the cap is shared, but the evidence itself is not held identical.
The Qwen2.5-7B experiment was post hoc; the Llama-3.1 and Qwen3.5
extensions were registered before their contexts were constructed,
after the initial result. On lookup-oriented TQA-Bench, the
Qwen2.5-7B gain is smaller (+1.3 points); effects vary across models.

\begin{table}[t]
\centering
\footnotesize
\setlength{\tabcolsep}{3pt}
\begin{tabular}{@{}lll@{}}
\toprule
LLM & BirdQA-R & DataBench-Test-L \\
\midrule
Qwen2.5-7B & $+10.5$ [5.6, 15.1] & $+11.9$ [7.4, 17.4] \\
Llama-3.1-8B & $+9.5$ [5.6, 13.8] & $+6.4$ [1.6, 10.9] \\
Qwen3.5-9B & $+6.2$ [2.0, 10.2] & $+14.1$ [10.0, 18.6] \\
\bottomrule
\end{tabular}
\caption{Shared-budget accuracy gains (points) and paired 95\% question-level intervals for Hybrid-2000 minus Coverage Alone.}
\label{tab:hybcov}
\end{table}

\subsection{Why Query-Aware Evidence Selection Matters}

\emph{Coverage Alone} applies 2K coverage to all questions.
\emph{Computation-First} selects a 600-token computation context under
broader eligibility rules, including grounded computation,
selective-anchor conjunctions, or answer-option matches; otherwise,
it uses a 1K coverage fallback. The latter is used on 0, 30, and
98 questions on the three main datasets. Computation-First can
include supporting rows and is not a computation-only input.

\begin{figure*}[t]
\centering
\includegraphics[width=0.95\textwidth]{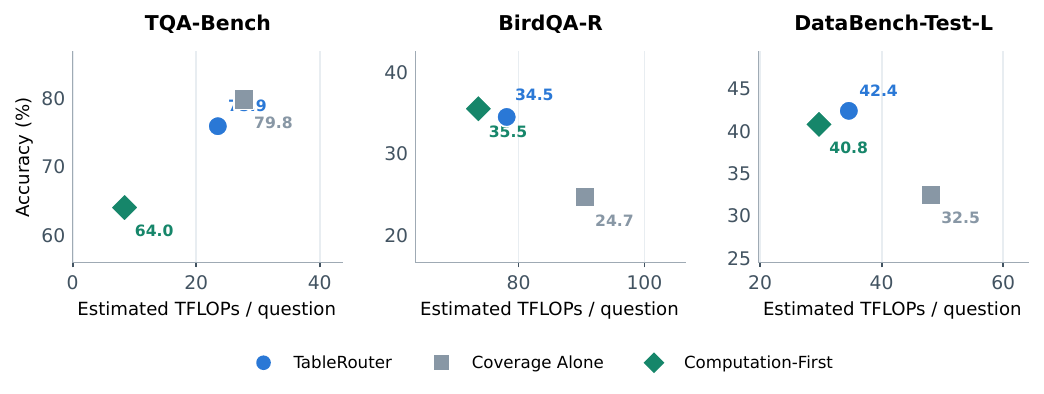}
\caption{Accuracy versus estimated per-question compute for the three main evidence policies with Qwen2.5-7B. Each panel is one benchmark; accuracy and cost are not normalized across datasets.}
\label{fig:j07}
\end{figure*}

Figure~\ref{fig:j07} visualizes the policy results already reported
in Table~\ref{tab:internal}. Coverage Alone is strongest on
lookup-oriented TQA-Bench; computation-oriented evidence performs
better on BirdQA-R; and \TR{} has the highest observed accuracy among
the three policies on DataBench-Test-L. Importantly, \TR{} stays
competitive across all three datasets with the \emph{same} routing
rule, while using less estimated compute than Coverage Alone in
each. This is a favorable operating point, not a claim that every
individual route decision is optimal.

\subsection{Post-Freeze Evidence beyond Development-Informed Benchmarks}

\begin{figure*}[t]
\centering
\includegraphics[width=0.92\textwidth]{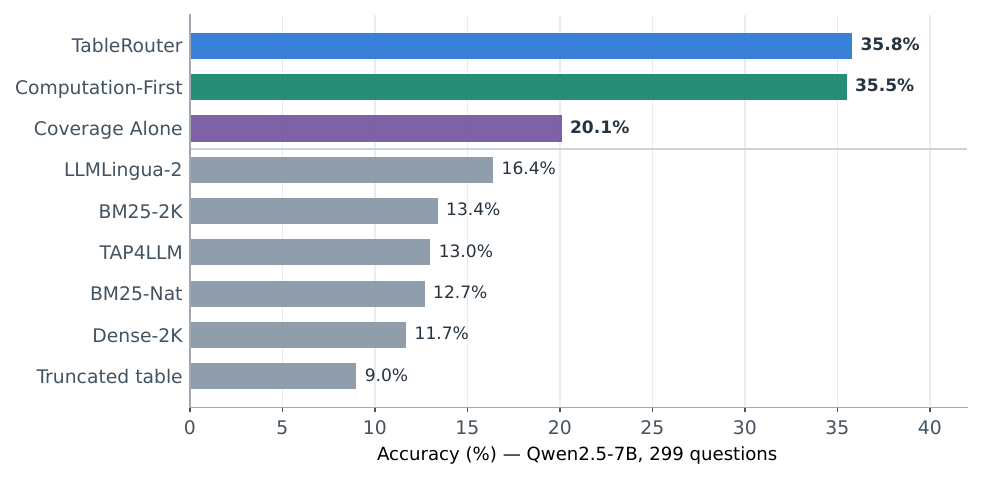}
\caption{Post-freeze BIRD accuracy with Qwen2.5-7B, covering 299 questions on 55 tables. The six external baselines are shown in gray; internal policy controls are distinguished from the primary \TR{} result.}
\label{fig:j08}
\end{figure*}

The post-freeze evaluation makes the system-level comparison on
previously unused databases especially informative. \TR{} answers
\textbf{107 of 299 questions (35.8\%)} correctly, against 49
(16.4\%) for LLMLingua-2, the strongest external baseline. The
+19.4-point difference has question-level interval [13.7, 25.8]
and table-level interval [9.2, 29.2]. Figure~\ref{fig:j08}
places this result beside the full external baseline group and
the two policy controls. Computation-First reaches 35.5\%, and its
0.3-point observed difference from \TR{} is inconclusive.
The primary gate chooses computation for 206 of 299 questions
(68.9\%), consistent with the computational character of many
SQL-derived questions; this route frequency alone does not establish
per-question correctness of the gate.

\section{Robustness and Question-Type Checks}
\label{app:robust}

\subsection{A Stronger Table-Augmentation Baseline}

The TAP4LLM+Stats control augments column types, counts, means,
minima, maxima, and distinct-value counts with a 2K evidence cap.
It is stronger than primary TAP4LLM in eight of nine model--dataset
settings, but does not produce question-filtered full-table statistics
by default. Table~\ref{tab:tapstats} shows that \TR{} retains higher
observed accuracy in all nine comparisons, by 4.8--29.7 points;
eight paired question-level intervals exclude zero. The exception
is Llama-3.1-8B on DataBench-Test-L, $+4.8$ [$-1.3$, $+11.6$].
For Qwen2.5-7B, a 2K TAP4LLM control without statistics scores
54.3\%, 16.4\%, and 27.0\%; adding these table-level statistics
changes the scores to 47.5\%, 16.4\%, and 35.4\%. This underscores
that budget changes and augmentation content are distinct effects.

\begin{table}[t]
\centering
\footnotesize
\setlength{\tabcolsep}{4pt}
\begin{tabular}{@{}llrrr@{}}
\toprule
Set & LLM & TAP+Stats & \TR{} & Gain \\
\midrule
TQA & Q2.5 & 47.5 & 75.9 & $+28.4$ \\
 & L3.1 & 52.8 & 76.6 & $+23.8$ \\
 & Q3.5 & 58.1 & 87.8 & $+29.7$ \\
Bird & Q2.5 & 16.4 & 34.5 & $+18.1$ \\
 & L3.1 & 14.8 & 30.9 & $+16.1$ \\
 & Q3.5 & 19.7 & 41.8 & $+22.0$ \\
DB-L & Q2.5 & 35.4 & 42.4 & $+7.1$ \\
 & L3.1 & 31.8 & 36.7 & $+4.8$ \\
 & Q3.5 & 29.9 & 47.3 & $+17.4$ \\
\bottomrule
\end{tabular}
\caption{Accuracy (\%) against TAP4LLM with column statistics and a 2K evidence cap. Gains are \TR{} minus the control, calculated before rounding.}
\label{tab:tapstats}
\end{table}

\subsection{Evidence Hints and Answer-Only Prompts}

BirdQA-R includes supplemental evidence hints for 272/304
questions. Removing them from every method leaves \TR{} at 29.3\%
accuracy with Qwen2.5-7B versus 14.8--18.8\% for the six rerun
primary baselines; the advantage does not depend on sharing these
hints. The computation-route frequency changes from 65.8\% with
hints to 55.9\% without them.

To check sensitivity to explanation length and output-token caps,
we use answer-only Qwen3.5-9B prompting. \TR{} reaches 37.2\%
on BirdQA-R and 50.8\% on DataBench-Test-L, exceeding the strongest
of truncation, BM25-Nat, and BM25-2K by 19.1 and 12.9 points,
respectively. The tested baseline results are 17.8\%/37.9\%
(truncation), 18.1\%/34.4\% (BM25-Nat), and 17.8\%/30.9\%
(BM25-2K). In a separate Llama template control, removing the
shared duplicated begin-of-text token gives \TR{} accuracies
81.2\%, 33.6\%, and 40.8\%, versus 73.3\%, 15.8\%, and
28.3\% for the tested baseline. These are selected post hoc
checks, not a second complete nine-setting evaluation.

\subsection{Generalization across Question Operations}

\begin{figure*}[t]
\centering
\includegraphics[width=0.94\textwidth]{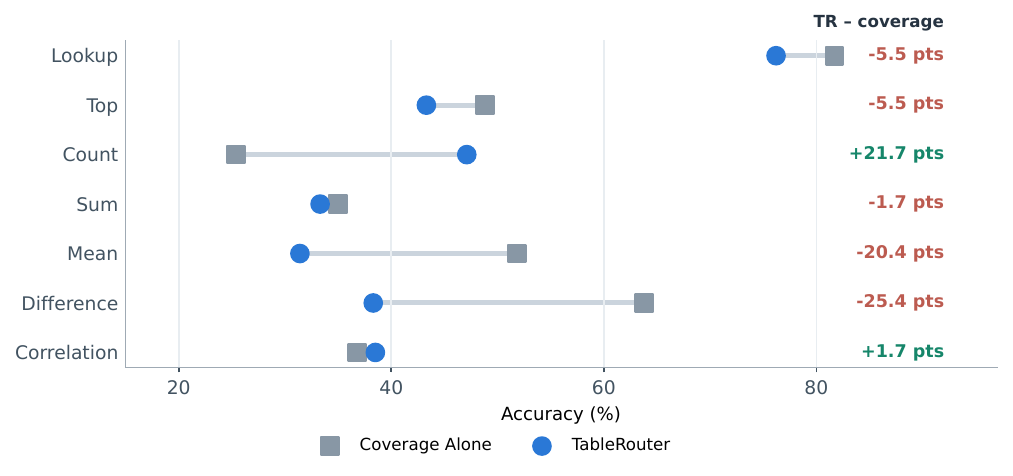}
\caption{Accuracy by operation on 1,660 mixed-type TQA-Bench questions, Qwen2.5-7B. Each pair compares \TR{} and Coverage Alone; labels give paired accuracy differences (points). The source tables are shared with the main TQA-Bench evaluation.}
\label{fig:j09}
\end{figure*}

A registered mixed-type TQA-Bench extension samples two questions
per table from six additional question types and adds 240 main-set
lookups. After excluding 20 all-NaN-option questions, the resulting
population contains 1,660 questions on the same 120 tables; it is
not a fresh-table test. Figure~\ref{fig:j09} contrasts the primary
\TR{} policy with Coverage Alone by operation. \textbf{Counts benefit
substantially from computed evidence}: 47.1\% accuracy for \TR{}
versus 25.4\% for Coverage Alone. By contrast, mean and difference
questions favor coverage, with gaps of 20.4 and 25.4 points.

This contrast helps separate computation from question interpretation:
a correct aggregate over a mismatched population or target cannot
answer the original question. Overall, \TR{} scores 44.2\% compared
with 49.0\% for Coverage Alone on this extension, a paired difference
of $-4.9$ [$-7.7$, $-2.2$] points. Excluding 70 questions matching an
earlier copy gives $-4.6$ [$-7.4$, $-1.7$]. These results identify
where better condition and operation matching would improve the gate;
they do not reverse the fixed-row evidence-use finding, where the
qualifying rows are held constant.

\section{Table Size and the Long-Table Regime}
\label{app:size}

\begin{figure*}[t]
\centering
\includegraphics[width=0.93\textwidth]{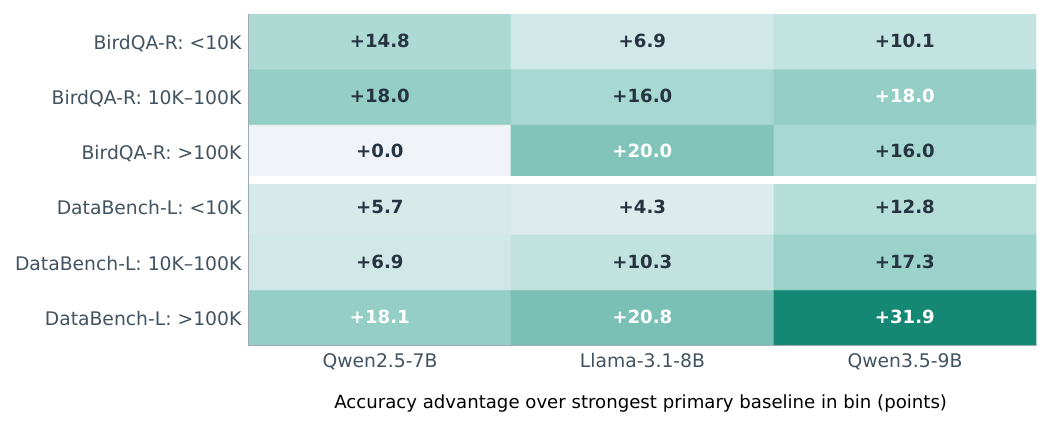}
\caption{Descriptive accuracy differences (points) between \TR{} and the strongest primary baseline within each source-table size bin and LLM. BirdQA-R and DataBench-Test-L subsets are shown; the zero cell is an observed tie, not a significance claim.}
\label{fig:j10}
\end{figure*}

The key evaluation constraint is that tables exceed a usable
\emph{context budget}, not one universal row threshold. The size-bin
analysis in Figure~\ref{fig:j10} is descriptive: bins were fixed
before final-limit runs but after earlier exploratory results, and
no within-bin significance test was registered. Each cell compares
\TR{} with the strongest of the six primary external baselines for
that model and bin, so the comparison baseline may differ by cell.

\textbf{The long-table advantage remains nonnegative in every shown
BirdQA-R and DataBench-Test-L bin.} On the two DataBench-Test-L
tables above 100,000 rows (72 questions), \TR{} reaches 51.4\%,
47.2\%, and 61.1\% across the three LLMs, versus 33.3\%, 26.4\%,
and 29.2\% for the strongest baselines within that subset.
Qwen2.5-7B prompt lengths across these table-size bins average
917--1,725 tokens, compared with 15,227--24,099 for truncation.

The BirdQA-R row bins contain 129 questions/29 tables below 10K
rows, 150/16 at 10K--100K, and 25/7 above 100K. DataBench-Test-L
has 210/6, 29/1, and 72/2 across the corresponding bins. The small
number of independent tables in some bins limits cross-table
inference. TQA-Bench is more appropriately grouped by table width;
its 20--50-column subset has two reversals favoring BM25-Nat. These
observations establish performance on large supplied tables, not
monotonic gains with increasing row count.

\section{Compute, Timing, and Index Reuse}
\label{app:compute}

\subsection{Cost Accounting and Decomposition}

The estimated per-question cost in Eq.~\eqref{eq:accounting}
charges table-index construction once and divides it over the
questions served by that table. Index terms include row embeddings
and clustering; inference terms include prefill and decoding.
Dataset averages are computed from individual questions rather
than by evaluating cost at the mean prompt length. These FLOPs
are analytical estimates, not energy or elapsed-time measurements.

\begin{figure*}[t]
\centering
\includegraphics[width=0.94\textwidth]{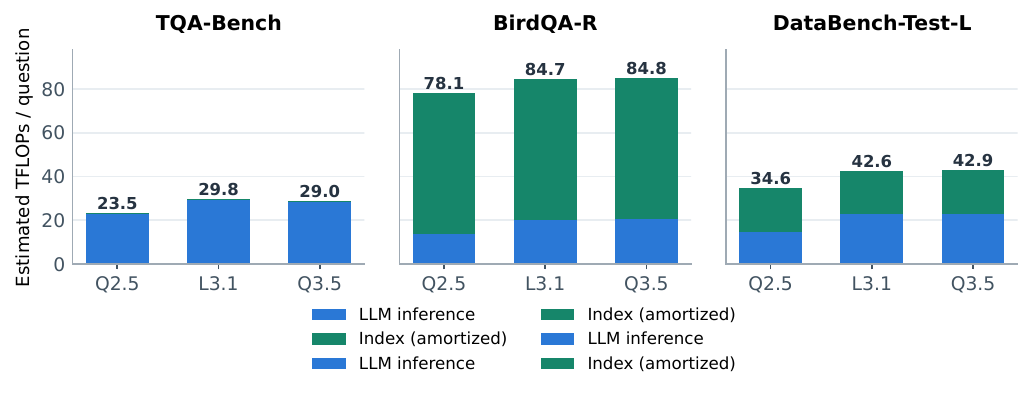}
\caption{\TR{} estimated per-question compute by benchmark and LLM: LLM inference plus amortized index construction. Labels are rounded totals (TFLOPs); small accounted online terms can cause differences from the sum of displayed components.}
\label{fig:j11}
\end{figure*}

Figure~\ref{fig:j11} and Table~\ref{tab:computedec} separate the
amortized index and LLM contributions. With Qwen2.5-7B, the indexed
share is about 82\% on BirdQA-R and 58\% on DataBench-Test-L.
Despite those build costs, \TR{} reduces estimated FLOPs by
77--95\% relative to truncated-table prompting across the nine
main settings when index cost is amortized over their questions.

\begin{table}[t]
\centering
\footnotesize
\setlength{\tabcolsep}{4pt}
\begin{tabular}{@{}llrrr@{}}
\toprule
Set & LLM & LLM TF & Index TF & Total TF \\
\midrule
TQA & Q2.5 & 22.9 & 0.6 & 23.5 \\
 & L3.1 & 29.1 & 0.6 & 29.8 \\
 & Q3.5 & 28.3 & 0.6 & 29.0 \\
Bird & Q2.5 & 13.7 & 64.4 & 78.1 \\
 & L3.1 & 20.3 & 64.4 & 84.7 \\
 & Q3.5 & 20.4 & 64.4 & 84.8 \\
DB-L & Q2.5 & 14.7 & 19.9 & 34.6 \\
 & L3.1 & 22.7 & 19.9 & 42.6 \\
 & Q3.5 & 23.0 & 19.9 & 42.9 \\
\bottomrule
\end{tabular}
\caption{Estimated TFLOPs per question for \TR{}, including amortized indexing. Entries are rounded independently.}
\label{tab:computedec}
\end{table}

For prompt length $L$, generation step $t$, $N$ parameters excluding
input embeddings and output projection, $n_\ell$ full-attention
layers, and attention width $d$, full-attention inference is
approximated by
\begin{equation}
\begin{aligned}
\mathcal F_{\rm pre}(L) &= 2NL+4n_\ell L^2d,\\
\mathcal F_{{\rm decode},t} &= 2N+4n_\ell(L+t)d.
\end{aligned}
\label{eq:flops}
\end{equation}
The output projection contributes $2Vh$ per generated token. The
Qwen3.5-9B model's eight full-attention and 24 Gated DeltaNet
layers are accounted for separately; the expression above is not
applied to all of them \citep{kaplan2020scaling,yang2025gated}.
The independent layerwise implementation check agrees at four
prompt lengths. Loaded BGE-small has 33,212,160 parameters excluding
its unused pooler, with 21.3M outside the input embeddings; row
encoding is capped at 512 tokens.

The accounting charges Dense-2K and TAP4LLM for row embeddings but
not clustering, and LLMLingua-2 for compressor computation. The
latter omits about 0.4\% attributable to per-chunk special tokens
and padding. CPU BM25 indexing is treated as free; the inherited
BM25-Nat budget construction is not charged the \TR{} build.
\TR{}'s online preprocessing is at most 0.0024 TFLOPs per question,
with at most another 0.002 TFLOPs from a coverage-scoring pass omitted.
On post-freeze BIRD, \TR{} totals 78.3 TFLOPs per question,
compared with 530.154 for truncation and 248.785 for BM25-Nat:
85.2\% and 68.5\% less, respectively, under these assumptions.

\subsection{Warm-Index Latency and Cold-Start Cost}

\begin{figure*}[t]
\centering
\includegraphics[width=0.94\textwidth]{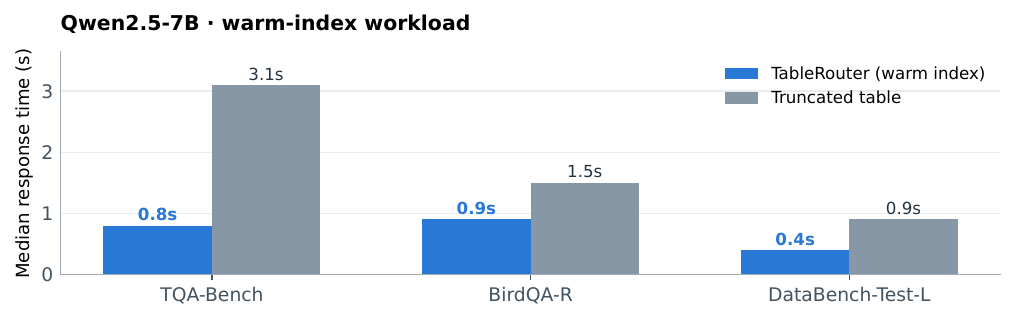}
\caption{Measured median warm-index response latency (seconds), Qwen2.5-7B on A100 80GB. Initial \TR{} index construction is excluded; the comparison concerns already indexed tables.}
\label{fig:j12}
\end{figure*}

Latency was measured for Qwen2.5-7B on an NVIDIA A100 80GB with
CPU clustering. The warm-index sample uses 60 questions per main
benchmark, with the source table and index loaded before timing.
Figure~\ref{fig:j12} shows median response times for online evidence
construction plus LLM generation: 0.8, 0.9, and 0.4 seconds for
\TR{} versus 3.1, 1.5, and 0.9 seconds for truncation. Reuse thus
reduces median latency on all three samples.

First-question measurements on the largest-build tables tell a
different story: \TR{} medians are 3.5, 815.0, and 858.9 seconds
across TQA-Bench, BirdQA-R, and DataBench-Test-L, compared with
3.9, 66.0, and 108.8 seconds for truncation. Building the largest
DataBench-Test-L index takes 834.5 seconds. The analysis therefore
supports the repeated-query workload rather than assuming a
cold-start advantage.

With $W_{\mathrm{ref}}$ and $W_{\mathrm{TR}}$ denoting per-question online
plus inference costs and with no charged shared-index cost for the
reference method, the analytical break-even question count is
\begin{equation}
\begin{aligned}
Q^*(T)&=\frac{\mathcal F_{\mathrm{index}}(T)}
{W_{\mathrm{ref}}-W_{\mathrm{TR}}},\\
& W_{\mathrm{ref}}>W_{\mathrm{TR}}.
\end{aligned}
\label{eq:payback}
\end{equation}
Using dataset-average online workloads, the largest BirdQA-R build
repays after 9.8--14.2 questions against truncation and 20.9--25.5
against BM25-Nat across LLMs. Actual thresholds depend on the
question workload; these are FLOP-based estimates, not time guarantees.

\section{Routing and Component Diagnostics}
\label{app:diag}

\subsection{Component Sensitivity}

Table~\ref{tab:components} examines individual interventions while
retaining the primary evidence pipeline where possible. A rows-only
policy is distinct from replacing row selection \emph{inside the
coverage branch}. The default model is evaluated as implemented,
without retroactively replacing it by a simplified configuration.

\begin{table*}[t]
\centering
\footnotesize
\setlength{\tabcolsep}{5pt}
\begin{tabular}{@{}l rr rr rr@{}}
\toprule
 & \multicolumn{2}{c}{TQA-Bench} & \multicolumn{2}{c}{BirdQA-R} & \multicolumn{2}{c}{DataBench-Test-L} \\
\cmidrule(lr){2-3}\cmidrule(lr){4-5}\cmidrule(l){6-7}
Configuration & Acc. & TF & Acc. & TF & Acc. & TF \\
\midrule
\TR{} & 75.9 & 23.5 & 34.5 & 78.1 & 42.4 & 34.6 \\
Without lexical grounding & 75.9 & 23.5 & 34.5 & 77.8 & 44.7 & 33.0 \\
Ranked rows in coverage branch & 75.9 & 23.3 & 33.6 & 78.2 & 41.2 & 34.7 \\
Ranked Rows Only & 76.6 & 27.4 & 18.1 & 90.5 & 26.7 & 48.2 \\
One prototype ($K=1$) & 77.0 & 23.5 & 35.2 & 77.8 & 43.4 & 34.6 \\
No anchor-score term & 74.0 & 23.5 & 33.9 & 78.4 & 42.1 & 34.7 \\
$K=1$ and ranked rows & 76.8 & 23.3 & 34.2 & 77.8 & 41.8 & 34.6 \\
\bottomrule
\end{tabular}
\caption{Qwen2.5-7B component controls: accuracy (\%) and estimated per-question TFLOPs. Ranked Rows Only is a separate complete policy; other row substitutions change the coverage branch.}
\label{tab:components}
\end{table*}

The rows-only policy sharply reduces accuracy on BirdQA-R and
DataBench-Test-L relative to \TR{}, reinforcing the importance of
combining retrieval with question-conditioned computation. Several
retrieval heuristics have no independently detectable accuracy effect
in these runs. Omitting grounding improves DataBench-Test-L by
2.3 points [0.3, 4.5], suggesting that lexical mismatches can reject
useful computations. A post hoc simplification with one prototype and
ranked-row coverage changes accuracy by less than one point on each
main benchmark, with zero-spanning intervals. This does not establish
component equivalence and is not substituted for the full system.
Across three index seeds, accuracy spreads are 0.7, 1.0, and
0.3 points, with unchanged route decisions.

\subsection{Where Computation and Coverage Recover Answers}

To understand why the primary gate can be useful across different
question populations, we compare the \emph{complete}
Computation-First and Coverage Alone policies on the subsets of
questions assigned to each \TR{} route. These are post hoc policy
comparisons, not component-level causal ablations. In the
computation-assigned subsets, Computation-First gains 30 net correct
answers over Coverage Alone on BirdQA-R and 31 on DataBench-Test-L.
On TQA-Bench's coverage-assigned questions, Coverage Alone yields
143 net additional correct answers over Computation-First. These
opposing effects explain the value of retaining both forms of
evidence, even though route selection remains imperfect.

\begin{figure*}[t]
\centering
\includegraphics[width=0.94\textwidth]{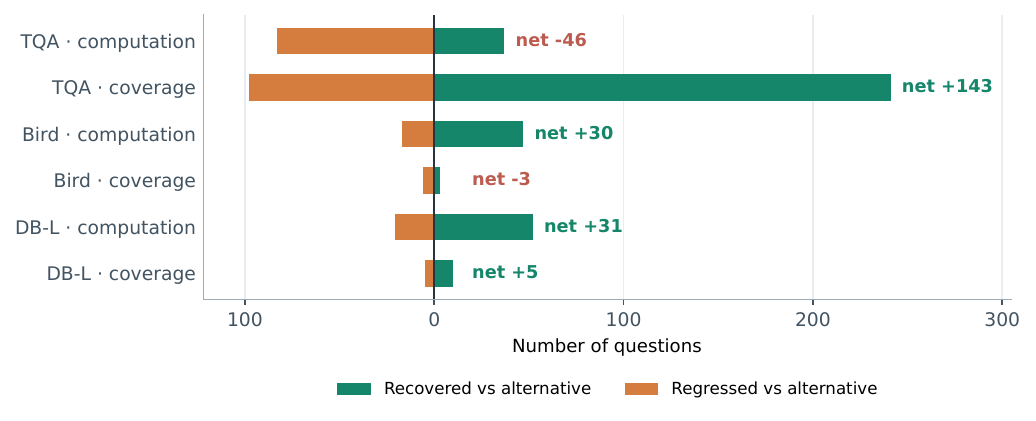}
\caption{Post hoc policy outcomes by the route selected by \TR{}. Bars show answers recovered and lost relative to the alternative complete evidence policy; labels give net differences. These are not estimates of a causal effect of an isolated component.}
\label{fig:j13}
\end{figure*}

The same decomposition also records 46 net losses from the
computation-oriented policy on TQA-Bench's computation-assigned
subset, alongside smaller coverage-subset differences of $-3$
(BirdQA-R) and $+5$ (DataBench-Test-L). The evidence therefore
supports question-dependent evidence selection while identifying
lookup interpretation as a meaningful refinement opportunity.
Additional mean/difference trace inspections distinguish errors in
selecting the qualifying population from cases where operands were
available but the final answer was incorrect; literal trace matches
alone do not establish semantic sufficiency or the LLM's internal
reasoning process.

\end{document}